%% file: main.tex
\documentclass[pdflatex,sn-basic]{sn-jnl}

\usepackage{graphicx}%
\usepackage{multirow}%
\usepackage{amsmath,amssymb,amsfonts}%
\usepackage{amsthm}%
\usepackage{mathrsfs}%
\usepackage[title]{appendix}%
\usepackage{xcolor}%
\usepackage{textcomp}%
\usepackage{manyfoot}%
\usepackage{booktabs}%
\usepackage{algorithm}%
\usepackage{listings}%

\usepackage{algorithmic}
\usepackage{array}
\usepackage[caption=false,font=normalsize,labelfont=sf,textfont=sf]{subfig}
\usepackage{stfloats}
\usepackage{url}
\usepackage{verbatim}
\usepackage{balance}

\usepackage{amssymb}
\usepackage{amsmath}

\usepackage{graphicx}
\usepackage{tabularx}
\usepackage{paralist}
\usepackage{enumitem}
\usepackage{amsthm}
\usepackage{colortbl}
\usepackage{multicol}
\usepackage{tikz}
\usepackage{float}
\usetikzlibrary{shapes.geometric, arrows}

\theoremstyle{thmstyleone}%

\theoremstyle{thmstyletwo}%

\theoremstyle{thmstylethree}%

\tikzstyle{startstop} = [rectangle, rounded corners, 
minimum width=5cm, 
minimum height=1cm,
text centered, 
draw=black, 
fill=red!30]

\tikzstyle{process} = [rectangle, rounded corners,
minimum width=4cm, 
minimum height=1cm, 
text centered, 
text width=2.5cm, 
draw=black, 
fill=orange!30]

\tikzstyle{fine_grained} = [rectangle, rounded corners,
minimum width=2.5cm, 
minimum height=1cm, 
text centered, 
text width=3cm, 
draw=black, 
fill=green!30]

\tikzstyle{reference} = [rectangle, 
minimum width=2.5cm, 
minimum height=1cm, 
text centered, 
text width=3.5cm, 
draw=black, fill=white!30]

\tikzstyle{arrow} = [thick,->,>=stealth]

\newcommand{\edit}[1]{#1}

\begin{document}

\title{Retrieval-Augmented Generation for Natural Language Processing: A Survey}





\author[1,2]{Shangyu Wu}
\author[2]{Ying Xiong}
\equalcont{Corresponding author.}
\author[3]{Yufei Cui}
\author[3]{Haolun Wu}
\author[3]{Can Chen}
\author[3]{Ye Yuan}
\author[1]{Lianming Huang}
\author[2]{Xue Liu}
\author[4]{Tei-Wei Kuo}
\author[1]{Nan Guan}
\author[2]{Chun Jason Xue}


\affil[1]{\orgname{City University of Hong Kong}}

\affil[2]{\orgname{Mohamed bin Zayed University of Artificial Intelligence}}

\affil[3]{\orgname{McGill University, Mila}}
\affil[4]{\orgname{National Taiwan University}}


\abstract{Large language models (LLMs) have achieved strong empirical performance in various fields, benefiting from their huge amount of parameters that store knowledge. 
However, LLMs still suffer from several key issues, such as hallucination problems, knowledge update issues, and lacking domain-specific expertise. 
The appearance of retrieval-augmented generation (RAG), which leverages an external knowledge base to augment LLMs, mitigates these limitations. 
This paper presents a systematic review of RAG techniques for natural language processing (NLP), with a focus on retrievers and retrieval fusions. 
We introduce a novel taxonomy of retrieval fusions, such as query-based, logits-based, latent, and parametric fusion, and provide structured comparisons across accessibility, efficiency, and use cases. 
The paper further examines RAG applications across diverse NLP tasks, discusses evaluation methodologies and benchmark limitations, and analyzes training paradigms with and without knowledge base updates. 
Finally, we explore industrial deployment considerations and identify emerging challenges and future directions, including security, efficiency, and graph-based retrieval.}

\keywords{Retrieval-augmented generation, natural language processing, vector database, large language model}



\maketitle

\input{1-introduction}
\input{2-Paradigm}
\input{3-retrievers}
\input{4-fusions}
\input{5-generators}
\input{8-tasks}
\input{7-benchmarking}
\input{6-training}

\input{9-applications}
\input{10-discussion}
\input{11-conclusion}
\input{12-declare}

\bibliography{ref}

\end{document}

%% file: 1-introduction.tex
\section{Introduction}

Large language models (LLMs)~\citep{llama, gemma, mistral, gpt4, 23iclr-glm} have achieved significant advancements in recent years and have become the cornerstone of various applications in the field of natural language processing (NLP).
These LLMs are typically pre-trained on a large amount of natural language corpus and then fine-tuned on the specific downstream tasks' datasets.
Recent works~\citep{19emnlp-lm-as-kb, 22arxiv-lmkb-survey, 22nips-rome, 24arxiv-lmkb-scale} demonstrate that the success of LLMs can be explained by the fact that LLMs implicitly store the learned knowledge in the parameters as internal memory and generate responses by retrieving answers from memory.
To store more knowledge for better generation performance, existing works generally enlarge the memory capacity by increasing the volume of parameters~\citep{22iclr-scaling-law, gpt3, 20arxiv-scaling-law, 22arxiv-scaling-law}.

Although existing LLMs have become a foundational component in modern NLP systems, several challenges still hinder the development of LLMs.
\edit{One of the most prominent challenges is the hallucination problem~\citep{23cs-hallucination-survey, 23acl-hallucination-mt, 23acl-hallucination-dialogue, 24ceur-hallucination},} which refers to the tendency of LLMs to generate responses that are coherent and fluent but factually incorrect.
Furthermore, the knowledge update issue poses a significant obstacle.
To update the knowledge stored in the LLMs' internal memory~\citep{22nips-rome, 23arxiv-editing-survey, 24arxiv-editing-survey}, it is necessary to retrain/fine-tune LLMs with new data, which is a costly process.
Additionally, general LLMs often lack domain-specific expertise~\citep{23emnlp-huatuogpt, 23nature-med-palm, 23arxiv-med-palm2, 24arxiv-saulm}. 
Building a domain-specific LLM demands substantial effort in data curation and model adaptation.

\edit{To address these challenges, recent works~\citep{20neurips-rag, 22icml-retro, 20icml-realm} have proposed augmenting LLMs with an external knowledge base (KB), known as Retrieval-Augmented Generation (RAG). 
By supplying the model with relevant, factually grounded information at inference time, RAG enables generation that is informed by verifiable sources rather than relying solely on parametric memory. 
This approach mitigates the hallucination problem by anchoring outputs in retrieved information. 
Furthermore, RAG addresses the issue of knowledge staleness by updating the external database allows the model to leverage current information without costly retraining. 
Similarly, domain-specific expertise can be imparted by constructing and integrating a specialized KB, effectively adapting a general-purpose LLM to specialized tasks. 
Consequently, RAG plays a pivotal role in enhancing the accuracy, adaptability, and reliability of LLMs across a broad spectrum of applications.}

\begin{table}[t]
\caption{Comparison of existing surveys [1]~\citep{22arxiv-rag-survey}, [2]~\citep{23arxiv-survey}, [3]~\citep{24arxiv-survey-ecust}, [4]~\citep{24arxiv-survey-pku}, [5]~\citep{24arxiv-survey-york}, and [6]~\citep{24arxiv-survey-polyu}. $\checkmark$ represents that the corresponding survey has introduced the content \textbf{in detail}. $\triangle$ indicates that the corresponding survey \textbf{briefly} introduced the content. $-$ represents that the corresponding survey \textbf{doesn't mention} the content.}
\label{tab:overview}
\centering
\begin{tabular}{lcccccccc} 
\toprule
\multicolumn{2}{c}{Modules} & [1] & [2] & [3] & [4] & [5] & [6] & Ours \\
\midrule
\multirow{2}{*}{Retriever} & Building 
& $-$          & $\checkmark$ & $\triangle$ & $\triangle$  & $\triangle$  & $\checkmark$ & $\checkmark$ \\
& Querying 
& $\triangle$  & $\checkmark$ & $\triangle$ & $\triangle$  & $\triangle$  & $\checkmark$ & $\checkmark$ \\
\multicolumn{2}{c}{Retrieval Fusions} 
& $\triangle$  & $\triangle$ & $\checkmark$ & $\checkmark$ & $\triangle$  & $\checkmark$ & $\checkmark$ \\
\multicolumn{2}{c}{Generator} 
& $\triangle$  & $\triangle$ & $\checkmark$ & $\checkmark$ & $\triangle$  & $\checkmark$ & $\checkmark$\\
\multicolumn{2}{c}{RAG on NLP tasks} 
& $\checkmark$& $\triangle$  & $\checkmark$ & $\triangle$  &$-$             & $\triangle$  & $\checkmark$ \\
\multicolumn{2}{c}{RAG Benchmarking} 
& $-$          & $\checkmark$ & $\checkmark$& $\checkmark$ & $\checkmark$ & $-$          & $\checkmark$ \\
\multicolumn{2}{c}{RAG Training/Update} 
& $-$          &  $\triangle$ & $\triangle$ & $\triangle$  & $\triangle$  & $\checkmark$ & $\checkmark$ \\ 
\multicolumn{2}{c}{RAG Tools} 
&$-$            & $-$         & $-$         & $\checkmark$ &$-$             & $\checkmark$ & $\checkmark$ \\
\multicolumn{2}{c}{Tutorial Codes} 
& $-$          & $-$         & $-$          & $-$          & $-$           & $-$          & $\checkmark$ \\
\bottomrule
\end{tabular}
\end{table}

\begin{figure*}[t]
\centering
\begin{tabular}{c}
\includegraphics[width=1.0\linewidth]{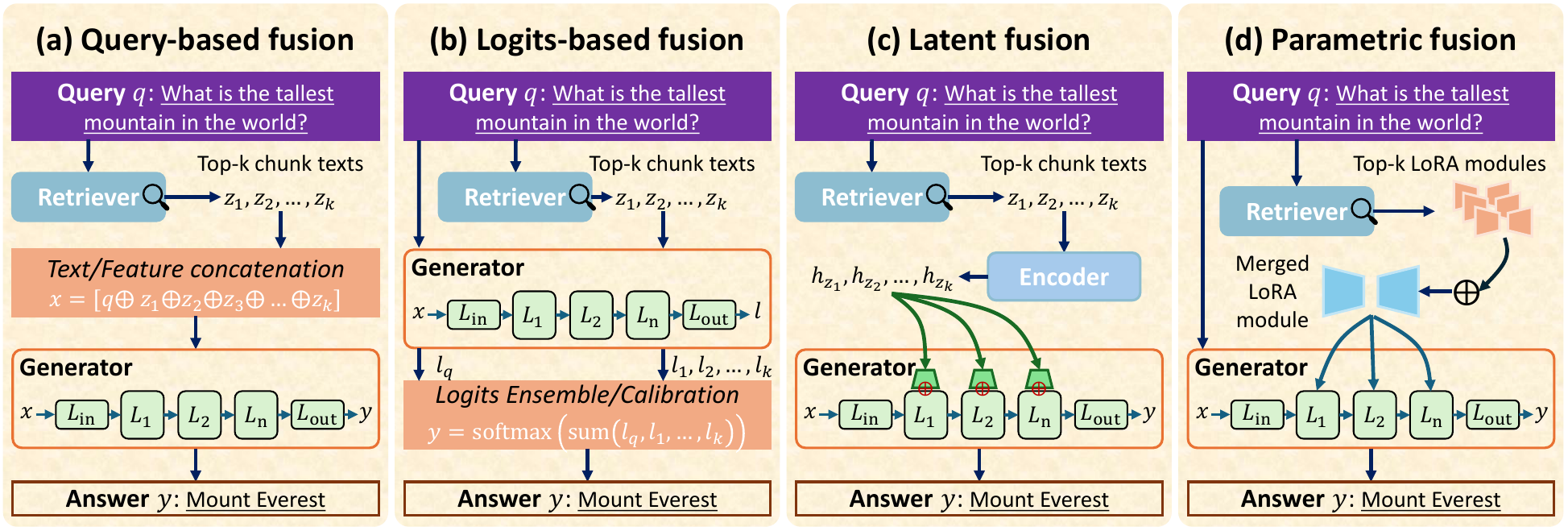} \\
\includegraphics[width=1.0\linewidth]{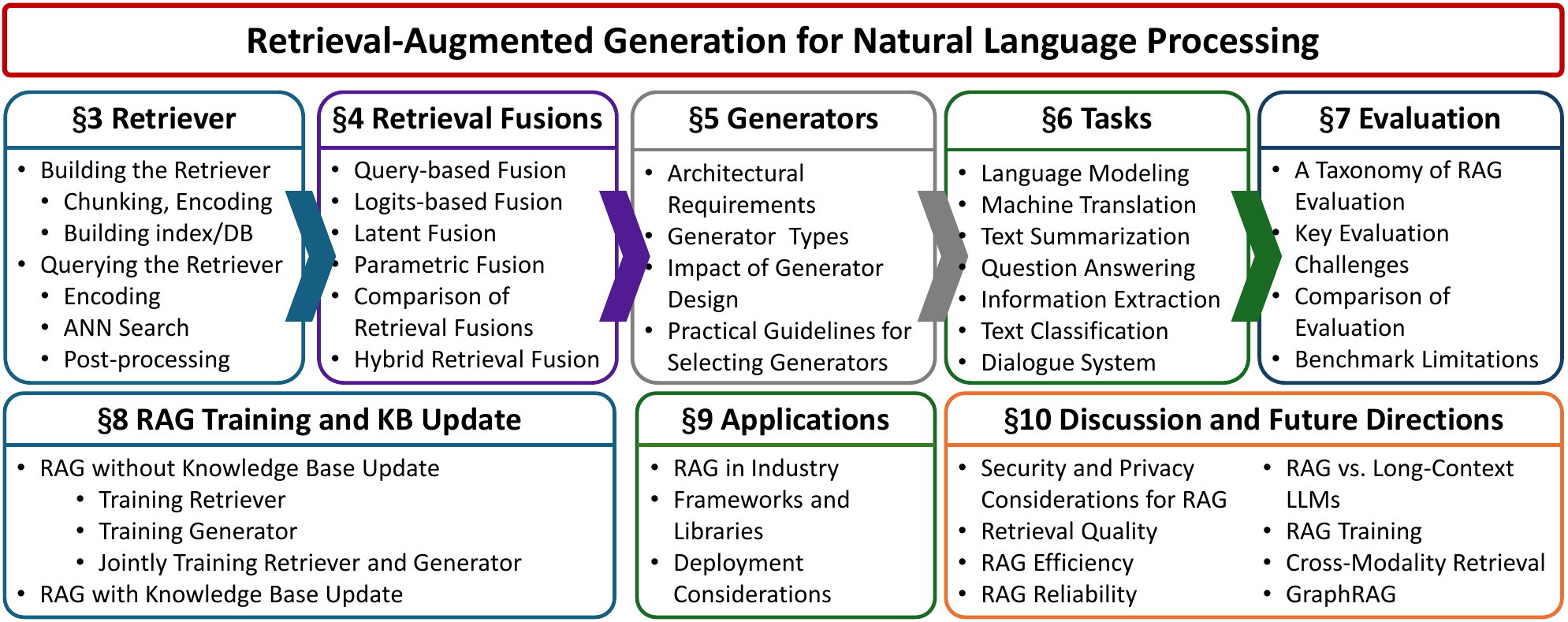}
\end{tabular}
\edit{\caption{The overview of retrieval-augmented generation for natural language processing. The inputs as queries are fed into both the retriever for retrieval knowledge and the generator for outputs. There are four kinds of retrieval fusions, including query-based fusion, logits-based fusion, latent fusion, and parametric fusion.}}
\label{fig:overview}
\end{figure*}

\textbf{Contributions.} 
\edit{This paper presents a systematic and comprehensive survey of RAG techniques for natural language processing. 
Although several prior surveys have explored this emerging paradigm~\citep{22arxiv-rag-survey, 23arxiv-survey, 24arxiv-survey-ecust, 24arxiv-survey-pku, 24arxiv-survey-york, 24arxiv-survey-polyu}, our work offers distinct perspectives and deeper technical insights, as summarized in Table~\ref{tab:overview}.}
The main contributions of this survey are as follows:
\edit{\begin{enumerate}
    \item We provide an end-to-end, system-oriented decomposition of RAG for NLP, complemented with algorithm-level walkthrough and time/space complexity analyses to bridge concepts and implementable pipelines\footnote{Tutorial codes are available on https://github.com/luffy06/RAG-Tutorials}.
    \item We propose a comprehensive taxonomy of retrieval fusion (query-, logits-, latent-, and parametric fusion) and a structured comparison across accessibility, efficiency, implementation complexity, and use cases, including practical insights on hybrid fusion designs.
    \item We synthesize RAG training and knowledge-update strategies under settings with and without datastore updates, linking retriever/generator optimization with parameter-efficient knowledge injection workflows.
    \item We comprehensively introduce the applications of RAG across a wide range of NLP tasks and RAG evaluation organizing metrics and methodologies.
    \item We also extend the survey toward practical scenarios with tooling/ecosystem and deployment considerations alongside emerging directions such as security/privacy or GraphRAG.
\end{enumerate}}

The structure of this paper is organized as follows and outlined in Figure~\ref{fig:overview}. 
Section~\ref{sec:para} provides an overview of RAG. 
Sections~\ref{sec:retriever} and~\ref{sec:fusion} then delve into the technical details of retrievers and retrieval fusion mechanisms, respectively. 
Section~\ref{sec:gen} examines the generator component. 
Subsequently, Section~\ref{sec:task} reviews the techniques employed in representative NLP tasks. 
Section~\ref{sec:eval} introduces evaluation methodologies and benchmarks for RAG systems.
Section~\ref{sec:training} discusses training paradigms for RAG, distinguishing between configurations with and without dynamic knowledge integration.
Section~\ref{sec:app} surveys RAG deployment in practical NLP scenarios. 
Section~\ref{sec:future} identifies promising directions for future research, and Section~\ref{sec:con} concludes the paper.

%% file: 2-paradigm.tex
\section{Overview}
\label{sec:para}

This section gives an overview of RAG for NLP. 
An RAG system utilizes the external knowledge base $\boldsymbol D $ to enhance the generation system. 
Taking external documents as an example, $\boldsymbol D $ consists of external documents, each of which contains a set of chunks $c_i \in \boldsymbol{C_i}$.
These chunks are transformed into vector embedding using an embedding model.
When inputting a query $q$, which is embedded as a vector, the retriever in the RAG system retrieves top-$k$ chunks $Z_q=\{z_1, z_2, \dots, z_k\}$ most relevant to the query $q$. The RAG system can use different retrieval fusion to fuse the retrieved chunks, which is discussed in Section \ref{sec:fusion}. The overall process is formulated as follows:
\begin{equation}
    \texttt{Retriever}(q, D) \rightarrow Z_q 
\end{equation}
\begin{equation}
     \texttt{Generator} (\texttt{Retrieval Fusion} (q, Z_q)) \rightarrow answer
\end{equation}
\edit{As shown in the above equations, the RAG system typically consists of three modules: the retriever, the generator, and the retrieval fusions.}

\textbf{Retriever} module usually comprises three components: 
an encoder for encoding inputs into embeddings, 
an efficient indexing that supports approximate nearest neighbor (ANN) search, 
and a vector database for storing external knowledge in the form of key-value pairs. 
The main challenge in the retriever module is \textit{finding the optimal trade-off between retrieval efficiency and retrieval quality}.
The retrieval efficiency refers to how fast the relevant information can be obtained, which involves accelerating encoding, efficient indexing, batch querying in the vector database, etc.
The retrieval quality refers to how relevant the information can be retrieved, which involves chunk representation learning, advanced ANN search algorithms, etc.

\textbf{Retrieval Fusions} aim to leverage the retrieved information to augment the generation.
\edit{These retrieval fusions can be categorized into four major types: query-based fusion, logits-based fusion, latent fusion, and parametric fusion.}
The query-based fusion augments inputs with retrievals before feeding them into the generators.
The logits-based fusion focuses on the output logits of generators and fuses the retrieval logits for more robust logits.
The latent fusion refers to introducing retrieval representations into the latent representations of generators, thus implicitly improving the models' performance.
\edit{The parametric fusion refers to the process of encoding external knowledge into Low-Rank Adaptation (LoRA) modules, which are subsequently merged into the model's weights to inject the retrieved information. Specifically, the top-$k$ most relevant LoRA modules are selected and integrated, thereby augmenting the model with task-specific knowledge without modifying its original parameters.}

\edit{\textbf{Generator} modules in RAG systems fall into three categories. Proprietary LLMs (e.g., GPT~\citep{gpt1, gpt2, gpt3, gpt4}, Gemini~\citep{gemini-1, gemma, gemini-1.5}) offer strong capabilities but cannot be deployed locally. Open-weight pretrained LLMs, such as Mistral~\citep{mistral}, Qwen~\citep{qwen2, qwen2.5}, DeepSeek~\citep{guo2025deepseek}, and Llama~\citep{llama, llama2, arxiv_llama3}, provide flexibility for local deployment. Retrieval-aware generators (e.g., RETRO~\citep{22icml-retro, 24icml-instructretro}, Enc-Dec~\citep{22neurips-encoder-decoder}) are specifically designed to incorporate retrieved information through dedicated fusion mechanisms.}

The workflow of the RAG system involves three steps: 
1) retrieving the relevant information from the external KB based on the given input query;
2) fusing the retrieved information with inputs or intermediate states based on the retrieval fusion; 
3) making predictions by generators based on the input query and corresponding retrievals.

%% file: 3-retrievers.tex
\section{Retriever}
\label{sec:retriever}

\begin{figure*}[t]
\centering
\includegraphics[width=1\linewidth]{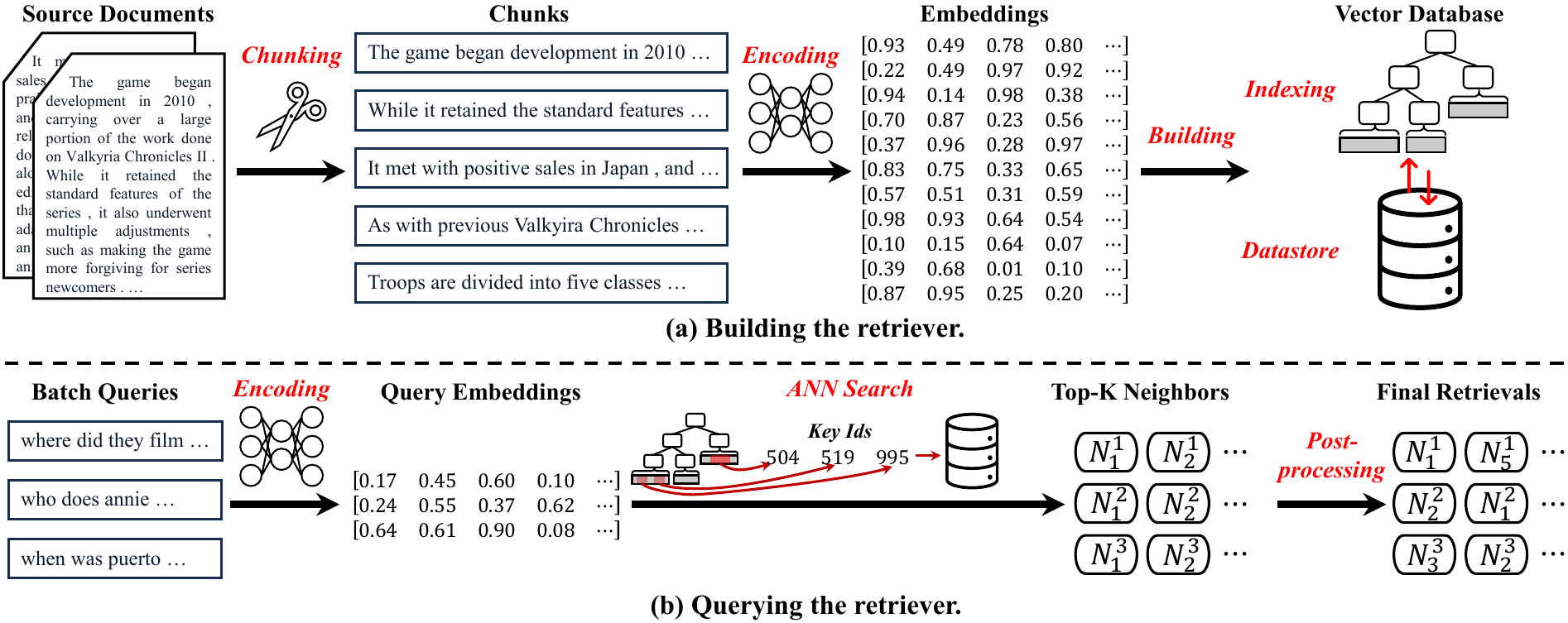}
\caption{Two stages of using the retriever.}
\label{fig:retriever}
\end{figure*}

Figure~\ref{fig:retriever} shows the two stages for using the retriever, which involve first building the retriever and then querying the retriever.
The following sections will introduce details about each stage.

\subsection{Building the Retriever}
This section will explain how to build a retriever using a large natural language corpus. 
As shown in Figure~\ref{fig:retriever} (a), the process involves three steps: chunking corpus, encoding chunks, and building the vector database.
Specifically, building the vector database includes building the ANN index and storing the data with key-value pairs.

\subsubsection{Chunking Corpus}
Chunking techniques generally refer to dividing large documents into small text chunks~\citep{16acl-sentence-chunking, 17acl-chunk, 20acl-rec-chunk, 22icml-retro, 22arxiv-sedr}, which is an indispensable key step in the process of building the retriever.
The intuitions behind chunking techniques are, 
(1) The texts or embeddings used for the indexing should be semantically independent, containing one core idea for models to encode. Short texts are more likely to be ambiguous, for example, the word ``apple`` can refer to a fruit or a company.
(2) Encoding a long sequence document would result in considerable resource overhead when using existing transformer-based models, while processing shorter text chunks can significantly accelerate the encoding process and save memory costs.
Therefore, the main challenge of the chunking techniques is to find the best chunking size to make a better trade-off between text semantics and encoding efficiency.

To solve the above challenge, three key points need to be considered when determining the chunking size:
\begin{compactenum}
    \item \textbf{Task's preference.} \edit{Different tasks benefit from different chunk sizes based on their information granularity. For example, for question answering tasks, they prefer short sentences, where the chunk size of 128 tokens is enough, as representative QA tasks only have 100-200 tokens at most per question~\citep{19tacl-nq,18emnlp-openbookqa}. However, for summarization tasks, they are usually long documents, where the choice for chunk size can be 512 tokens due to lengthy documents~\citep{18emnlp-xsum}.}
    \item \textbf{Encoder's preference.} Different encoder models have varying encoding capabilities on texts with different lengths. For example, models in the sentence-transformer~\citep{sen-trans} behave better on a single sentence, while the text-embedding-ada-002~\citep{text-emb-ada} is good at longer texts. \edit{The choice of chunk size usually depends on the max input sequence length of the encoder model, e.g., 512 tokens for BERT-based models.}

    \item \textbf{Query's preference.} The length of the user's queries should be aligned with the chunking size, which implicitly aligns the amount of contextual information in chunks with that in queries, thus improving the relevance between queries and retrievals. For example, a retrieval database built on short phrases may be useless for queries with long documents.
\end{compactenum}
Overall, there is no golden rule for determining the chunking size, and it depends on the specific RAG scenarios.

There are basically three types of chunking techniques, including \textbf{the chunking with fixed length}, \textbf{the semantic chunking}, and \textbf{the content-based chunking}. 
Chunking with a fixed length is the simplest way to split documents sequentially using a length hyperparameter.
The semantic chunking cuts documents based on semantics, such as the period character or the newline character that represents the end of the sentence. 
Existing state-of-the-art natural language processing toolkits, such as NLTK~\citep{nltk} and spaCy~\citep{spacy}, have provided convenient sentence-cutting methods. 
The content-based chunking segments documents according to the unique structural characteristics. 
For example, electronic medical records can be easily segmented based on the sections, or programming codes can be segmented based on function blocks.

\edit{Before chunking and encoding, text preprocessing (e.g., normalization, removing boilerplate/HTML artifacts, deduplication, and consistent handling of casing/punctuation) can influence the quality of chunk representations. 
Prior studies~\citep{24is-preprocess} show that preprocessing choices can measurably affect downstream performance for both Transformer-based models and traditional models, suggesting that corpus preparation should be treated as an important knob when optimizing retrieval quality and robustness in RAG.}

\subsubsection{Encoding Chunks}
\label{sec:encoding}
Encoding refers to numericalizing textual chunks as vector representations (embeddings). 
These embeddings generally capture the semantics of the chunks, enabling the retriever to perform similarity searches based on content relevance rather than just keyword matching.

According to the sparsity of the embeddings, there are two kinds of encoding methods, i.e., \textbf{sparse encoding} and \textbf{dense encoding}.
The sparse encoding represents text by creating high-dimensional vectors where most elements are zero.
The basic sparse encoding is one-hot encoding~\citep{one-hot}, which represents a word with a high-dimensional vector as large as the vocabulary table size but only marks the value corresponding to the presence of the word as one.
The embeddings produced by such encodings are called the one-hot vector.
Other common sparse encodings include:
\begin{enumerate}

    \item \textbf{Bag of Words (BoW)}~\citep{bow}. This encoding improves one-hot encoding by replacing the zero-one counting with the frequency counting. However, BoW ignores the syntax and word order in the documents and focuses on statistical information, thus only expressing limited semantics.
    \item \textbf{Term Frequency-Inverse Document Frequency (TF-IDF)}~\citep{tf-idf}. This encoding not only counts the occurrence (frequency) of each word but also adjusts these counts based on how common the word is across all documents (inverse document frequency). TF-IDF helps emphasize words that are more descriptive of the document's content.
    \item \textbf{BM25}~\citep{bm25} is a probabilistic ranking algorithm used in information retrieval to estimate the relevance of documents to a search query by balancing term frequency, inverse document frequency, and document length normalization, ensuring robust scoring even for long or short documents. BM25 focuses on lexical matches and is computationally efficient, making it a cornerstone of traditional search engines. 
\end{enumerate}
Sparse encoding is an efficient way to encode textual chunks.
However, such encoding methods may not capture deeper semantic meanings well.

The dense encoding generates vectors where each dimension can capture a range of semantic features, and most elements are non-zero floating points.
The dense embeddings are generally produced by (deep) neural network models, 
\begin{compactenum}
    \item \textbf{BERT~\citep{19naacl-bert} and Variants.} Bidirectional Encoder Representation from Transformers (BERT) is a typical pre-trained transformer model, generating dense semantic embeddings that capture the contextual information. Other BERT variants, such as RoBERTa~\citep{19arxiv-roberta}, DistilBERT~\citep{19arxiv-distilbert}, and ELECTRA~\citep{20iclr-electra}, further improve the semantic representations with advanced learning techniques. 
    \item \textbf{Siamese Encoders.} This is a type of neural network designed to learn the similarity between inputs, which is usually trained with contrastive learning. Existing state-of-the-art siamese encoders are DPR~\citep{20emnlp-dpr}, SimCSE~\citep{21emnlp-simcse}, Contriever~\citep{22_tmlr_contriever}.
    \item \textbf{LLM-based Encoders.} This type of encoder benefits from the powerful representation capability of LLMs. LLMs, which contain billions of parameters and are pre-trained on vast amounts of data covering a wide range of topics, have advanced semantic language understanding capabilities. Typical LLM-based encoders are text-embedding-ada-002~\citep{text-emb-ada}, bge-embedding~\citep{bge-embedding}, mxbai-embedding~\citep{mxbai-embedding}, MedCPT~\citep{23bib_medcpt}.
\end{compactenum}
Compared to sparse encoding, dense encoding leverages deep neural networks, especially transformers~\citep{17nips-attention}, to capture broader linguistic and semantic information.
Dense encoding is widely used in most representation scenarios.
Moreover, some works also utilized hybrid methods to encode text for leveraging both lexical and semantic information~\citep{22naacl_drboost, 23acl_elife}.
\subsubsection{Building the Index}
\label{building index}
Indexing in the vector database aims to accelerate the search process for data similar to high-dimensional query embedding.
Unlike common indexing in databases, indexing in the vector database mainly focuses on supporting efficient ANN search~\citep{21tbd-faiss, 24arxiv-faiss, 20icml-scann, 23tkde-anns-survey, 24tkde-quant-anns} rather than transaction operations like insertion, deletion, and update.
The key challenge of indexing is making a good trade-off between search quality and search efficiency.
To solve the challenge, there are various specific optimizations in both algorithmic aspects and systematic aspects to be explored, including choices of similarity metrics, dimension reduction on embeddings, advanced ANN indexing, system-level optimizations, hardware-aware optimization, and so on.
Here we focus on the optimizations that most strongly affect search quality and efficiency.

\textbf{Choice of Similarity Metrics.} 
The similarity metrics are the basic components in the retriever, which measure the degree of relevance between query embeddings and chunk embeddings.
The similarity metrics would affect the search quality. 
Typical similarity metrics include cosine similarity, Euclidean similarity, and Manhattan distance.

\textbf{Dimension Reduction on Embeddings.} 
Reducing the dimensionality of embeddings can improve search efficiency but at the risk of harming the semantic representations. 
The basic but effective dimension reduction is the principal component analysis (PCA). 
The PCA is a simple statistical technique that transforms the original data into a new coordinate system while retaining the most important features. 
Another popular and advanced dimension reduction is locality-sensitive hashing (LSH). 
LSH significantly reduces the dimensionality by mapping the data into buckets but preserves the similarity of the original input data.
The intuition behind LSH is that the nearest neighbors will be mapped into the same buckets.
Unlike LSH, product quantization (PQ)~\citep{11tpami-pq} is another popular and effective DR technique for ANN search. 
The core idea of the PQ is to divide the high-dimensional space into smaller, independently quantized subspaces.
Each subspace creates a codebook of different quantized integers to form the representative and compact vectors.
The above techniques enable efficient storage and fast approximate search but may lose semantic information.
Recent work~\citep{23emnlp-autocompressor} proposed a new technique named AutoCompressor that reduces the dimension of embeddings by compressing the original context into semantically shorter embeddings.

\textbf{Advanced ANN Indexing.}
ANN Indexing generally refers to the methods or structures used to organize and manage data so that the approximate-nearest-neighbor search process is optimized for retrieval quality and retrieval efficiency.
This paper will introduce several advanced ANN indexing techniques.
\begin{enumerate}

    \item \textbf{The InVerted File system with Product Quantization (IVFPQ)} \citep{24arxiv-faiss} is a simple but effective indexing framework that combines two powerful techniques to enable an efficient and scalable ANN search process. The main idea of IVFPQ is first to cluster the data for coarse-grained partition and then to compress the data within each cluster into sub-vectors for fine-grained quantization. The coarse-grained clustering (the IVF component) significantly reduces the search space, while the fine-grained quantization (the PQ component) ensures a high retrieval performance.
    \item \textbf{The Hierarchical Navigable Small World (HNSW)}~\citep{hnsw} uses a hierarchical graph structure to perform ANN search in high-dimensional spaces efficiently. Specifically, HNSW treats high-dimensional vectors as nodes and connects them with their nearest neighbors. The multi-layer graph structure is determined probabilistically to ensure fewer nodes at higher layers for efficient search.
    \item \textbf{Tree-based Indexing} aims to organize high-dimensional vectors in tree-liked structures, such as KD-Trees~\citep{19kdd-kd-tree}, Ball Trees~\citep{23icde-ball-tree} and VP-Trees~\citep{15prl-vp-tree}. Typical tree-based indexing is Approximate Nearest Neighbors Oh Yeah (Annoy)~\citep{annoy}, which uses a forest of trees built based on random projections to separate the vector space into multiple hyperplanes for efficient ANN search. 
\end{enumerate}

\subsubsection{Building the KB with Key-Value Pairs}

The vector database used in the KB is a specialized database that stores and manages data as a collection of key-value pairs, where keys are the unique identifiers of high-dimensional embeddings and values are the domain-specific knowledge. 
Since the amount of data stored in the KB may be quite large, the storage engine, such as LMDB~\citep{lmdb} or RocksDB~\citep{rocksdb}, should be capable of efficient retrieval and data persistence.
A key design decision in the knowledge base is what should be stored as values.
For example, for question-answer tasks, when adding retrievals to prompts, the naive but effective way is to store the question embedding as the key and question-answer pairs as the value.
This can help the generation process as retrievals are used as demonstrations for models.
Recent works have proposed various state-of-the-art vector databases, including the indexing and the knowledge base, such as Milvus~\citep{21sigmod-milvus, 22vldb-manu}, FAISS~\citep{24arxiv-faiss, 21tbd-faiss},  LlamaIndex~\citep{22github-llamaindex}, etc.


\begin{algorithm}[t]
\caption{\small Building the retriever.}
\label{alg:build}
\begin{algorithmic}[1]
\small
\REQUIRE A natural language corpus $D=\{d_1,\ldots, d_n\}$ for building the KB, an encoder $\mathcal{E}$ for encoding chunks.\\
\ENSURE The index $\mathcal{I}$ and the key-value store $\mathcal{S}$.
\STATE $\mathcal{K}=\{\}, \mathcal{V}=\{\}$; 
\FOR{$d_i \in D$}
    \STATE $c^1_i, \ldots, c^m_i = Chunk(d_i)$; \COMMENT{Split each data $d_i$}
    \FOR{$j$ from $1$ to $m$}
        \STATE $e^j_i = \mathcal{E}(c^j_i)$; \COMMENT{Encode each chunk $c^j_i$}
        \STATE Add $e^j_i$ into $\mathcal{K}$ and $c^j_i+c^{j+1}_i$ into $\mathcal{V}$; \COMMENT{Take next chunk as an example}
        \STATE The $\mathcal{K}$ and $\mathcal{V}$ persist in the storage (e.g., SSD) if necessary;
    \ENDFOR
\ENDFOR
\STATE Build the index $\mathcal{I}$ with $\mathcal{K}$;
\STATE Store $\mathcal{K}$ and $\mathcal{V}$ into the key-value store $\mathcal{S}$;
\RETURN $\mathcal{I}$ and $\mathcal{S}$;
\end{algorithmic}
\end{algorithm}

\subsubsection{Code Demonstrations}

Algorithm~\ref{alg:build} shows detailed steps to build the retriever.
Lines 2-8 present the chunking and the encoding process for a natural language corpus containing multiple documents.
In line 6, algorithm~\ref{alg:build} takes the concatenation of the current chunk and the next chunk as the value.
Notably, the choice of value can vary for different tasks.
Another practical issue is that the memory cost of all keys and values may exceed the memory capacity of the server in a practical scenario.
Thus, it is recommended that the keys and values persist in the storage if necessary.
\edit{
The time complexity is $O(n\cdot (C_{\text{chunk}}+m\cdot C_{\text{encode}})+C_{\text{build\_index}})$, where $C_{\text{chunk}}$, $C_{\text{encode}}$, and $C_{\text{build\_index}}$ are the cost of chunking one document, encoding one chunk, and building the index for all chunks.
The space complexity is $O(M\cdot d)$, where $M$ is the number of all chunks and $d$ is the feature dimension of the encoder.
}

\subsection{Querying the Retriever}

This section will explain how to query the pre-built retriever, which basically includes three steps as shown in Figure~\ref{fig:retriever}(b): encoding queries, ANN search, and post-processing.

\subsubsection{Encoding Queries and ANN Search}
To align with the pre-built embedding space, the retriever uses the same encoder to encode queries during the querying stage.
The ANN search leverages the pre-built indexing and vector database to find similar data via ANN searching algorithms and then retrieves the corresponding values.

\textbf{Searching the index} refers to searching the pre-built index, finding the top-k nearest neighbors, and returning the unique identifiers of the k nearest neighbors. 
The nearest neighbor search process depends on indexing algorithms or structures. 
Taking IVFPQ as an example, the search process first compares the query embedding with cluster embeddings and selects several candidate clusters for further search. 
Then, within each cluster, the search process performs the same product quantization on the query embedding and finds the top-k nearest neighbors based on the distance. 
Finally, the search process merges all nearest neighbor candidates and re-orders all candidates for the final top-k nearest neighbors.

\textbf{Retrieving values from vector database} fetches the corresponding values based on the nearest key identifiers. 

\begin{algorithm}[t]
\caption{\small Query the retriever.}
\label{alg:query}
\begin{algorithmic}[1]
\small
\REQUIRE A query input $q$, an encoder $\mathcal{E}$ for encoding chunks, the index $\mathcal{I}$, the key-value store $\mathcal{S}$, the parameter $k$.\\
\ENSURE Top-$k$ nearest neighbor knowledge.
\STATE $e=\mathcal{E}(q)$; 
\STATE $\{idx_1, \ldots, idx_k\} = \mathcal{I}.Search(e, k)$; \COMMENT{Search the top-$k$ nearest neighbors}
\STATE $\{v_1, \ldots, v_k\}=\mathcal{S}.Fetch(\{idx_1, \ldots, idx_k\})$; \COMMENT{Fetch the values of the neighbors}
\STATE $\{v_{j_1}, \ldots, v_{j_k}\}=PostProcess(\{v_1, \ldots, v_k\})$
\RETURN $\{v_{j_1}, \ldots, v_{j_k}\}$;
\end{algorithmic}
\end{algorithm}

\subsubsection{Post-Processing}

The post-processing involves a set of techniques after the initial retrieval step.
These techniques aim to refine, enhance, or adapt the retrievals based on the specific task objectives.
This section will list some typical post-processing techniques.

\textbf{Reranking} aims to reorder the retrieved information based on task-specific objectives. 
The intuition is that the knowledge is retrieved based on task-agnostic metrics, such as Euclidean distance. 
Existing reranking methods~\citep{23acl-ear, 20acl-retrieve-edit-rerank, 22arxiv-ialm, 23arxiv-refreshllms} mostly design different architectures or strategies to reorder the retrieved information.

\subsubsection{Code Demonstrations}

After building the retriever, this section demonstrates the detailed steps of querying the retriever to obtain the top-$k$ nearest neighbor knowledge in algorithm~\ref{alg:query}, including encoding the query (line 1), performing the approximate nearest neighbor search (line 2), and fetching the knowledge for fusion (line 3).
These three steps depend on the specific APIs of encoders, indexing, and the vector database.
After obtaining the top-$k$ retrievals, optimizations for post-processing are applied (line 4).
\edit{
The time complexity is $O(C_{\text{encode}} + C_{\text{ann}} + k\cdot C_{\text{IO}}+C_{\text{post}})$, where $C_{\text{ann}}$, $C_{\text{IO}}$, and $C_{\text{post}}$ are the cost of performing ANN search, fetching neighbor values from the disk, and post-processing.
The space complexity is $O(k\cdot d)$.
}

%% file: 4-fusions.tex
\section{Retrieval Fusions}
\label{sec:fusion}

Retrieval fusions refer to how to leverage the retrieved information to improve generators' performance.
Basically, there are \edit{four} types of retrieval fusions: query-based fusions, logits-based fusions, latent fusions, \edit{and parametric fusions}.

\begin{algorithm}[t]
\caption{\small Query-based Fusions.}
\label{alg:query-fusion}
\begin{algorithmic}[1]
\small
\REQUIRE A query input $q$, top-$k$ nearest neighbor knowledge $\{v_1, \ldots, v_k\}$, an encoder $\mathcal{E}_f$ and a decoder $\mathcal{D}_f$ for feature concatenation, the generator $\mathcal{G}$ for text concatenation.\\
\ENSURE Generated response $y$.
\IF{Use the text concatenation}
\STATE $x=v_1 \oplus \ldots \oplus v_k \oplus q$; \COMMENT{Concatenate neighbor texts and query}
\STATE $y=\mathcal{G}(x)$;
\ELSE
\STATE $e_q=\mathcal{E}_f(q), e_{v_j}=\mathcal{E}_f(v_j), j \in \{1, \ldots, k\}$;
\STATE $e_x=e_q \oplus e_{v_1} \oplus \ldots \oplus e_{v_k}$; \COMMENT{Concatenate embeddings of neighbors and query}
\STATE $y=\mathcal{D}_f(e_x)$
\ENDIF
\RETURN $y$;
\end{algorithmic}
\end{algorithm}

\subsection{Query-based Fusion}
The simplest and most direct retrieval fusion is query-based fusion, which integrates the retrieved information with input queries to generate responses.
The query-based fusion can be further categorized into two sub-classes according to the type of concatenated information, i.e., text concatenation and feature concatenation.

Text concatenation involves performing query-based fusion with raw texts, making it particularly suitable for proprietary LLMs like GPT-4. 
These models function as black-box systems with limited interaction capabilities, typically offering only API access to users.
Existing works~\citep{20icml-realm, 20neurips-rag, 23arxiv-ralm} directly concatenate the input with the top-$k$ retrieved sentences/documents to form the query for generators.
To better use the in-context learning capability of LLMs, some works~\citep{20acl-templated-retrieval, 22acl-reina, 23acl-udr, 23arxiv-refreshllms} design effective prompt templates to integrate retrieved information and inputs.
To address the issue of lengthy inputs after concatenating retrievals, recent studies~\citep{23arxiv-dil, 24iclr-recomp, 23arxiv-leancontext, 23arxiv-filco, 23emnlp-tcrallm, chen2026refilter} have introduced methods for assigning importance weights to elements within the retrieved information base and filtering out less relevant contexts based on these weights.

The feature concatenation involves merging the encoded retrievals with the input features.
A simple yet effective approach is FID~\citep{21eacl-fid}, which first encodes the retrieved passages into sparse or dense representations and then takes the concatenated features as the input for a generator.
The state-of-the-art performance of the FID demonstrates the efficacy of feature concatenation.
The follow-up works~\citep{21neurips-emdr2, 23acl-pgra, 23icml-lumen, 23jmlr-atlas, 23acl-recap} further improve the FID by jointly tuning the retriever and the encoder, thereby enhancing the representations of retrieved information.
Besides, Chen et al.~\citep{22neurips-retroprompt} concatenate the representations of related knowledge as demonstrations for prompt learning, yielding better generalization.

Algorithm~\ref{alg:query-fusion} presents how to leverage query-based fusions to fuse retrieved information.
For those using text concatenation~\citep{20icml-realm, 23arxiv-ralm}, algorithm~\ref{alg:query-fusion} first concatenates the retrieved texts and inputs (line 2), then feeds the concatenated input into the generator.
Notably, since there is a limit to the maximum input length of existing language models, concatenating too many retrievals would result in a truncation of the concatenated input, which may cut the given input.
Therefore, designing the prompt template is the key step for this branch of work.
For those using feature concatenation~\citep{21eacl-fid, 23acl-pgra}, algorithm~\ref{alg:query-fusion} first leverages an encoder to obtain the feature (line 5), then concatenates the feature of input and retrievals (line 6), finally passes the concatenated feature into a decoder model (line 7).
\edit{
The time complexity for text concatenation is $O(\mathcal{G}(L_q+\sum_{i=1}^k L_i))$, where $L_q, L_1, \ldots, L_k$ are the sequence length of query and retrievals.
If the generator is based on Transformer, then the time complexity is $O((L_q+\sum^k_{i=1}L_i)^2)$.
The time complexity for feature concatenation is $O(k\cdot C_{\text{encode}}+\mathcal{D}_f(L_q+\sum^k_{i=1}L_i))$.
The space complexity for text concatenation is $O(L_q+\sum^k_{i=1}L_i)$, for feature concatenation is $O((L_q+\sum^k_{i=1}L_i)\cdot d)$.
}

\begin{algorithm}[t]
\caption{\small Logits-based Fusions.}
\label{alg:logits-fusion}
\begin{algorithmic}[1]
\small
\REQUIRE A query input $q$, top-$k$ nearest neighbor knowledge $\{v_1, \ldots, v_k\}$, the generator $\mathcal{G}$.\\
\ENSURE Generated response $y$.
\STATE $y_q=\mathcal{G}(q)$;
\FOR{$j$ from $1$ to $k$}
\STATE $y_{v_j}=\mathcal{G}(v_j)$
\ENDFOR
\IF{Use ensemble}
\STATE $y=\lambda \sum_j y_{v_j} + (1-\lambda) y_q$;
\ELSE
\STATE $\lambda_t=Calibrate(y_q, y_{v_1}, \ldots, y_{v_k})$
\STATE $y=\lambda_t \sum_j y_{v_j} + (1-\lambda_t) y_q$;
\ENDIF
\RETURN $y$;
\end{algorithmic}
\end{algorithm}

\subsection{Logits-based Fusion}
The logits-based fusion refers to incorporating the retrieved information into the output layers.
Basically, retrieved information would be fed into the same model to obtain the logits for enhancing or calibrating the predictions.
Therefore, logits-based fusion can be categorized into two branches, i.e., ensemble-based fusion and calibration-based fusion.

Ensemble-based fusion treats the logits from the retrieved information as part of an ensemble of predictions.
Such ensemble-based fusion can significantly improve the generalization and robustness of the model~\citep{23emnlp-eara, 20iclr-knn-lm, 21iclr-knn-mt}.
One notable work of ensemble-based fusion is kNN-LM~\citep{20iclr-knn-lm}, which aggregates the logits of the top-$k$ nearest neighbors' targets and then interpolates the final predictions.
Similar to kNN-LM, Khandelwal et al.~\citep{21iclr-knn-mt} propose kNN-MT to enhance the machine translation using retrievals' logits, which is also followed by a branch of works~\citep{21acl-ijcnlp-adaptive-knn-mt, 23arxiv-knn-adapter}.

Different from ensemble-based fusion, calibration-based fusion uses the logits from the retrieved information as a form of calibration for the model's predictions.
Specifically, Jiang et al.~\citep{22emnlp-robust-knn-mt} propose a confidence-enhanced kNN-MT that refines the kNN distribution and interpolation weights with the neural machine translation confidence.
Li et al.~\citep{23emnlp-source-context} propose to leverage the source context to calibrate the retrieval-augmented neural machine translation.

Algorithm~\ref{alg:logits-fusion} demonstrates the detailed steps of using the logits-based fusion to integrate the retrieved information.
This branch of work first treats retrievals as similar data to augment the model (lines 2-4).
For ensemble, algorithm~\ref{alg:logits-fusion} leverages a hyperparameter to fuse the retrieval logits and the output logits (line 6).
For calibration, algorithm~\ref{alg:logits-fusion} dynamically determines the parameter based on the retrieval logits and the output logits (line 8).
Then, algorithm~\ref{alg:logits-fusion} performs the same fusion with the computed parameter (line 9).
\edit{
The time complexity is $O((k+1)\cdot \mathcal{G}(L))$ and the space complexity is $O((k+1)\cdot V)$, where $L$ represents any sequence length of query input or retrievals, $V$ is the vocabulary size.
}

\begin{algorithm}[t]
\caption{\small Latent Fusions.}
\label{alg:latent-fusion}
\begin{algorithmic}[1]
\small
\REQUIRE A query input $q$, top-$k$ nearest neighbors $\{v_1, \ldots, v_k\}$, the encoder $\mathcal{E}$, the generator $\mathcal{G}$ containing $l$ pairs of modules $\{(\mathcal{M}^A_1, \mathcal{M}^F_1), \ldots\}$, where $\mathcal{M}^A_i$ and $\mathcal{M}^F_i$ are the attention module and the FFN module at layer $i$, $\mathcal{M}^C_i$ is the cross-attention module used in attention-based latent fusions.\\
\ENSURE Generated response $y$.
\IF{Use the attention}
\STATE $h^F_0 = q$;
\FOR{$i$ from $1$ to $l$}
\STATE $h^A_i=\mathcal{M}^A_i(h^F_{i-1})$;
\STATE $e_{v_1}, \ldots, e_{v_k}=\mathcal{E}(v_1, \ldots, v_k, h^A_i)$
\STATE $h^R_i=\mathcal{M}^C_i(h^A_i, e_{v_1}, \ldots, e_{v_k})$; \COMMENT{Cross-attention}
\STATE $h^F_i=\mathcal{M}^F_i(h^R_i)$
\ENDFOR
\STATE $y=LM\_HEAD(h^F_l)$
\ELSE
\STATE $e_{v_1}, \ldots, e_{v_k}=\mathcal{E}(v_1, \ldots, v_k)$
\STATE $h^F_0 = q$;
\FOR{$i$ from $1$ to $l$}
\STATE $h^A_i=\mathcal{M}^A_i(h^F_{i-1})$;
\STATE $h^R_i=h^A_i+\frac{1}{k}\sum_j w_j e_{v_j}$ \COMMENT{Weighted sum}
\STATE $h^F_i=\mathcal{M}^F_i(h^R_i)$
\ENDFOR
\STATE $y=LM\_HEAD(h^F_l)$
\ENDIF
\RETURN $y$;
\end{algorithmic}
\end{algorithm}

\subsection{Latent Fusion}
\label{sec:latent}
The latent fusion investigates merging the retrieved information into the hidden states of generators for a better generation. 
Based on the introduction method, latent fusion can be further classified into two categories: attention-based and weighted-addition.

One notable contribution of attention-based fusion is the RETRO~\citep{22icml-retro}. 
RETRO represents a pioneering effort in pre-training retrieval-based LLMs, introducing a new cross-attention module to integrate retrieved information directly into the model's hidden states. 
A significant finding from this work is demonstrating a scaling law for the retrieval database, where RETRO, with a 2 trillion token database, attains performance comparable to that of major models like GPT-3 and Jurassic-1, albeit with 25 times fewer parameters. 
Customizing the transformer model in RETRO highlights the potential of pre-trained, retrieval-enhanced architectures in improving the efficiency and scalability of LLMs.

In addition to RETRO, other studies~\citep{21acl-monolingual-mem, 22iclr-mem-transformer, 22iclr-tome, 22neurips-encoder-decoder, 23iclr-retmol} have contributed to the field by leveraging new attention modules to introduce external knowledge.
Typically, Li et al.~\citep{22neurips-encoder-decoder} have extended the RETRO model by decoupling the context encoding from the model inference. 
Wu et al.~\citep{22iclr-mem-transformer}, Wang et al.~\citep{23neurips-longmem} store the hidden attention keys and values into external memory and retrieve the knowledge from the memory using an attention mechanism.

Due to the high complexity of the attention mechanism, another branch of work adopts lightweight (weighted) additions to introduce retrieved information.
Fevry et al.~\citep{20emnlp-eae} propose the EAE model that retrieves top-$k$ related entities' embeddings from a learnable external memory and adds entities' embeddings to the hidden states of the model.
Wu et al.~\citep{24iclr-refusion} propose ReFusion, which explores various learnable reranking schemes to first re-weight the embeddings of the retrieved information and then use weighted addition to incorporate them into the hidden states of the model.
Those approaches signify a growing trend towards models that dynamically select and integrate relevant information, paving the way for more sophisticated and nuanced language generation and understanding.

Algorithm~\ref{alg:latent-fusion} shows the steps of using latent fusion to introduce the retrieved information into the hidden states of the generator.
For attention-based latent fusion, algorithm~\ref{alg:latent-fusion} first encodes the retrievals with the output states of the attention module (line 5), then uses a cross-attention module to fuse the retrieval features into the hidden state (line 6).
Different from attention-based latent fusion, weighted-addition-based latent fusion adopts a more lightweight way to incorporate retrieved information (lines 10-19).
Algorithm~\ref{alg:latent-fusion} first encodes the retrievals before feeding them into the generator (line 11), which can be done offline and directly stored as values in the knowledge base.
Then, algorithm~\ref{alg:latent-fusion} learns a set of weights to add the retrieval features on the hidden states of generators (line 15).
\edit{
The time complexity for cross-attention module is similar to that of the attention module, but for weighted addition module is $O(k\cdot d)$.
The space complexity for both module is similar, i.e., $O(k\cdot d)$, storing the feature vectors of retrievals.
}

\edit{\subsection{Parametric Fusion}}

\edit{
\begin{algorithm}[t]
\caption{\small Parametric Fusions.}
\label{alg:para-fusion}
\begin{algorithmic}[1]
\small
\REQUIRE A query input $q$, top-$k$ nearest neighbors $\{v_1, \ldots, v_k\}$, the generator $\mathcal{G}$ with weight $w$.\\
\ENSURE Generated response $y$.
\STATE Get the top-$k$ LoRA-style modules $\{\Delta w_1, \ldots, \Delta w_k\}$ based on $\{v_1, \ldots, v_k\}$;
\STATE Merge those LoRA modules into one module $\Delta w=f(\Delta w_1, \ldots, \Delta w_k)$;
\STATE Update the LoRA modules into the generator $\hat w=w+\Delta w$;
\STATE $y=\mathcal{G}_{\hat w}(q)$;
\STATE Remove the LoRA modules for the next query $w=\hat w-\Delta w$;
\RETURN $y$;
\end{algorithmic}
\end{algorithm}
}

\edit{
Parametric fusion refers to injecting retrieved information into the parameters of the generator, rather than fusing them in hidden states. 
The core idea is to represent each document (or passage) as a lightweight parametric representation (e.g., LoRA style update~\citep{22iclr-lora, 23nips-qlora}) that can be plugged into the generator’s feed-forward networks (FFN). 
In practice, this paradigm usually involves an offline stage that converts documents into parameter-efficient ``knowledge modules'', and an online stage that follows a Retrieve–Update–Generate workflow: the system retrieves top-$k$ relevant documents, merges (or composes) their corresponding parameter updates to obtain a temporary adapted model, and then generates answers using the updated parameters. 
}

\edit{
Several recent works explore this direction with different design choices on how to obtain and fuse document-level parameters. 
PRAG~\citep{25sigir-prag} is a representative framework that parameterizes each document into LoRA-style updates offline and, at inference, merges the retrieved document-specific updates into a single plug-in module for answering. 
DyPRAG~\citep{25arxiv-dyprag} replaces per-document LoRA modules with a lightweight parameter translator that maps documents to parametric updates on the fly.
This enables test-time dynamic generation of LoRA modules and reduces the storage costs of parametric retrievals.
Poly-PRAG~\citep{26arxiv-poly-prag} shifts from the one-document–one-adapter paradigm to a many-to-few encoding. 
Poly-PRAG only learns a limited number of LoRA adapters and uses a document-level routing mechanism to select adapters for each document;
Then, the activated adapters are merged by routing weights.
This can further improve scalability and storage costs.
Such parametric fusions don't modify the architecture of base generators and maintain the same efficiency as the original inference.
}

\edit{
Algorithm~\ref{alg:para-fusion} demonstrates a generic parametric fusion pipeline. 
Given a query, the system retrieves top-$k$ documents, then obtains their parametric representations either by loading precomputed adapters (e.g., PRAG) or by translating documents into adapters on-the-fly (e.g., DyPRAG/PolyPRAG) (Line 1).
Then, the system performs parameter merging, such as summation, routing-weighted merging, or orthogonal merging, to obtain the final LoRA updates (Line 2).
Next, the system merges the LoRA weights to the generator weights and generate the final answer (Lines 3-4).
Finally, the system recovers the model weights for the next query (Line 5).
Since after merging, the inference process is exactly the same as that of the original LLMs, the time complexity is the same.
The space complexity is $O(N_w\cdot l\cdot d \cdot r)$, where $N_w$ is the number of weight matrices and $r$ is the rank of LoRA.
}

\subsection{Comparison of Different Retrieval Fusions}

\edit{
\begin{table}[t]
\caption{Comparison of different retrieval fusions.}
\label{tab:comp}
\centering
\begin{tabular}{p{2cm}p{2cm}p{2cm}p{1cm}p{1cm}p{1.5cm}p{2.5cm}}
\toprule
\textbf{Retrieval Fusions} & \textbf{Representative Works} & \textbf{Accessibility} & \textbf{Memory} & \textbf{Latency} & \textbf{Impl. Compl.} & \textbf{Scenarios} \\
\midrule
Query-based (Text Concat) 
& REALM, RAG, REINA, RALM 
& Black-box, White-box 
& High & High & Low
& Rapid deployment, frequently updated knowledge, proprietary LLMs \\
\midrule
Query-based (Feature Concat) 
& FID, RETRO-PROMPT, LUMEN 
& White-box (require fine-tuning decoders) 
& High & Moderate & Moderate
& Knowledge-intensive tasks, encoder-decoder architecture \\
\midrule
Logits-based (Ensemble) 
& kNN-LM, kNN-MT, kNN-Adapter 
& Black-box, White-box (require logits)
& Moderate & Moderate & Moderate
& Lightweight integration, plug-and-play \\
\midrule
Logits-based (Calibration) 
& Robust-kNN-MT, Source-Context 
& Black-box, White-box (require logits)
& Moderate & Moderate & Moderate
& Robust generation, uncertainty calibration, domain adaptation \\
\midrule
Latent (Attention) 
& RETRO, Enc-Dec, LONGMEM 
& White-box (require architecture modification) 
& Low & Moderate & High
& Large-scale pre-training, complex reasoning tasks\\
\midrule
Latent (Weighted Addition) 
& EAE, ReFusion 
& White-box (require architecture modification)
& Low & Low & High
& Entity enhancement, lightweight injection, and friendly fine-tuning \\
\midrule
Parametric 
& PRAG, DyPRAG, Poly-PRAG 
& White-box (require parameters injection)
& Low & Low & High
& Static knowledge, domain knowledge adaptation \\
\bottomrule
\end{tabular}
\end{table}
}

\edit{
To facilitate a systematic comparison, we summarize the key characteristics of the retrieval fusions discussed above in Table~\ref{tab:comp}, covering their accessibility, memory footprint, latency, and typical application scenarios. 
Each fusion method exhibits distinct advantages and limitations, and the choice of method depends on the specific deployment constraints and task requirements.
}


\edit{
\textbf{Accessibility.} Only query-based fusion (text concatenation) and logits-based fusion are compatible with black-box LLMs, as they only require input text or output logits, both commonly exposed via APIs. 
In contrast, other fusions demand white-box access, requiring architectural modifications (e.g., cross-attention modules) or parameter injection (e.g., LoRA adapters).
Those retrieval fusions usually require fine-tuning or even pre-training the LLMs.
}


\edit{
\textbf{Memory and Latency.} 
Parametric fusion achieves the highest memory and inference efficiency: after merging the retrieved LoRA module, the model behaves exactly like the base LLM, with all knowledge embedded in the weights, incurring no extra cost. 
Latent fusion also exhibits low memory overhead—it only stores compact representations of retrieved information (e.g., k d-dimensional vectors, where d is smaller than the LLM’s hidden size). 
However, attention‑based latent fusion introduces moderate latency due to cross‑attention computation, while weighted‑addition variants remain efficient. 
Logits‑based methods strike a middle ground: they require running the LLM k times (once per neighbor) to obtain logits, which can be batched to reduce latency at the cost of increased peak memory. 
Query‑based methods are the least efficient; text concatenation dramatically lengthens the input sequence, inflating the key-value (KV) cache and attention time. 
Feature concatenation improves throughput by encoding each retrieval independently, but still suffers from high memory usage due to long sequence length after the concatenation.
}


\edit{
\textbf{Implementation Complexity. }
Query-based text concatenation is the simplest (low complexity), as it operates entirely outside the model, i.e., retrieved texts are merely prepended to the input prompt, requiring no knowledge of the LLM's internal architecture. 
Feature concatenation and logits-based methods (both ensemble and calibration) entail moderate complexity: they interact with the model only at specific interfaces, e.g., after the encoder or at the output layer, without needing to modify internal layer structures. 
In contrast, latent fusion (attention and weighted addition) and parametric fusion demand high implementation complexity. 
These methods require deep architectural understanding, often involving modifications such as inserting cross-attention modules, introducing learnable reranking mechanisms, or injecting LoRA adapters. 
Moreover, they typically necessitate fine-tuning or even pre-training to achieve effective integration, making them suitable only for scenarios where full model access and substantial development resources are available.
}

\edit{
\textbf{Applicable Scenarios.} 
Query-based fusion with text concatenation enables rapid deployment with frequently updated knowledge, especially suitable for proprietary LLMs. 
Feature concatenation targets knowledge-intensive tasks, specifically for encoder-decoder architectures.
Logits-based ensemble methods offer plug-and-play integration, while calibrated variants enhance robustness under retrieval noise. 
attention-based latent fusion is particularly suitable for complex reasoning and pre-training; 
weighted addition provides lightweight injection for entity-centric tasks. 
Parametric fusion is best for static domain knowledge, encoding expertise into reusable adapters.
}

\edit{\subsection{Hybrid Retrieval Fusion}}

\edit{
Beyond using a single fusion strategy, recent works have explored hybrid fusions that combine multiple retrieval fusions to leverage their complementary strengths.
The most common paradigm is to use query-based fusion to inject retrieved information into the prompt, and then augment the inference with another fusion strategy (e.g., logits-based fusion, latent fusion, or parametric fusion) to improve robustness and information utilization. 
Text concatenation is often chosen as the ``base'' because it only modifies the input without requiring access to hidden states/logits or additional model training. 
For example, DVD~\citep{24emnlp-dvd} combines query-based fusion (text concatenation) and logits-based fusion (ensemble) to improve the generation quality. 
And PRAG~\citep{25sigir-prag} proposes to use the parametric fusion to inject knowledge, but further improves performance by integrating the query-based fusion (text concatenation).
}

\edit{
However, hybrid fusion is not limited to text concatenation. 
The text concatenation always introduces high computational and memory overheads due to the long input sequence.
Therefore, exploring other combinations of different retrieval fusions becomes a necessity.
For example, integrating logits‑based fusion with latent fusion can inject the knowledge into models via lightweight modules, without increasing input sequence length and enabling deep contextual reasoning.
With the development, we anticipate a richer landscape of hybrid fusions, where different strategies are adaptively selected or blended based on task requirements and resource constraints.
}

%% file: 5-generators.tex
\section{Generators}
\label{sec:gen}

\edit{In a RAG system, the \emph{generator} is the conditional language model that maps the user query $q$ and retrieved information $Z_q=\{z_i\}_{i=1}^k$ to an output sequence $y$.
Compared to a standalone LLM, a RAG generator must not only be fluent, but also (i) consume longer and noisier contexts, (ii) use retrieved information instead of relying on parametric memory, and (iii) ideally ground responses with verifiable attribution.
Most modern generators are Transformer-based models~\citep{17nips-attention} and are dominated by decoder-only LLMs such as GPT-series~\citep{gpt1, gpt2, gpt3, gpt4}, Llama-series~\citep{llama, llama2, arxiv_llama3}, Gemini/Gemma~\citep{gemini-1,gemini-1.5,gemma}, DeepSeek~\citep{guo2025deepseek}, Qwen series~\citep{qwen2, qwen2.5}, and Mistral/Mixtral~\citep{mistral,24arxiv_moe}.
These generators typically incorporate architectural optimizations such as RMSNorm~\citep{19nips-rmsnorm}, rotary position embedding (RoPE)~\citep{24neurocomputing-rope}, and grouped-query attention (GQA)~\citep{23emnlp-gqa}, which are especially relevant in RAG because retrieval may increase the input length and amplify inference-time memory/latency costs.}

\edit{\subsection{Architectural Requirements for Effective RAG}
A generator is ``RAG-effective'' only when its architecture and training make it both capable of ingesting retrieved information and willing to use it.
We summarize key requirements and their architectural implications:}
\edit{\begin{itemize}
    \item \textbf{Information Capacity and Context Management.}
    Query-based fusion (e.g., text concatenation that concatenates retrieved information into the prompt) requires large context windows and efficient use of attention/KV-cache usage.
    Recent LLMs extend context length substantially (e.g., Llama~3 up to 128K~\citep{arxiv_llama3}, DeepSeek-V2 supports 128K~\citep{guo2025deepseek}, Gemini~1.5 reports million-token-scale context~\citep{gemini-1.5}).
    However, long context does not guarantee robust information usage. Models can be position-sensitive and degrade when relevant information is ``lost in the middle''~\citep{23arxiv-lost-in-middle}.
    Therefore, generator design must consider not only supporting long-context generation but also effective context utilization under long inputs.
    \item \textbf{Utilizing Information beyond Simple Concatenation.}
    When substantial information is retrieved, naïve query-based fusion often causes attention dilution and budget waste.
    Architectures that \emph{separate information encoding from generation} can scale better.
    RETRO~\citep{22icml-retro} modifies the architecture of naive Transformer blocks by introducing a cross-attention module between the self-attention module and FFN module to inject the encoded information into the model.
    This line of works~\citep{22neurips-encoder-decoder,22icml-retro} exhibit extremely strong language modeling capability by pre-training the model.
    ReFusion~\citep{24iclr-refusion} also proposes to introduce some lightweight modules and fine-tunes the modified model for robustness.
    Those works demonstrate that architectural fusions can further improve the models' performance but at the cost of training.
    \item \textbf{Grounding, Attribution, and Controllability.}
    RAG applications often require citations, abstention when information is missing, and controllable behaviors (e.g., ``answer-only-from-information'').
    This typically needs either retrieval-aware training objectives or explicit control tokens and supervision.
    For instance, Self-RAG trains a single LM to retrieve on-demand and to generate/refine with special reflection tokens, improving factuality and citation behavior~\citep{23arxiv-self-rag}.
    \item \textbf{Robustness to Noisy/Conflicting Retrieval.}
    Retrieval quality is sometimes imperfect; retrieved information can be irrelevant or contradictory.
    Generators benefit from mechanisms that reduce over-reliance on any single passage, such as document-level marginalization (as in RAG's latent document variable)~\citep{20neurips-rag}, filter unrelated retrieval~\citep{chen2026refilter}, or training-time exposure to mixed-quality retrieval (e.g., Atlas, Self-RAG)~\citep{23jmlr-atlas,23arxiv-self-rag}.
    \item \textbf{Efficiency under Retrieval-Augmented Inference.}
    RAG often increases latency due to the retrieval and the increasing generation cost with longer prompts.
    Thus, inference-optimized attention designs are practically important:
    GQA reduces KV-cache overhead while preserving quality~\citep{23emnlp-gqa}; DeepSeek-V2 compresses KV cache via Multi-head Latent Attention (MLA) and adopts MoE for economical scaling~\citep{guo2025deepseek}.
\end{itemize}}

\edit{\subsection{Generator Types}
From an architectural and deployment standpoint, generators in RAG can be categorized more informatively than ``open vs. closed'':
\begin{itemize}
    \item \textbf{Proprietary/API LLMs.}
    Examples include GPT-4~\citep{gpt4} and Gemini~\citep{gemini-1,gemini-1.5}.
    They are attractive for strong general capability and engineering maturity, but usually constrain RAG to prompt-time fusion (concatenation, instruction templates, and multi-call orchestration).
    The key limitation is architectural immutability. One cannot insert cross-attention information modules or do retrieval-aware pretraining.
    Nevertheless, black-box RAG can still improve by optimizing retrieval and prompt composition.
    RePlug shows that one can treat the LLM as frozen and tune the retriever using LLM signals, improving black-box LLMs without changing generator weights~\citep{23arix-replug}.
    \item \textbf{Open-Weight Decoder-only LLMs.}
    Models such as Llama~\citep{llama,llama2, arxiv_llama3}, DeepSeek~\citep{guo2025deepseek}, enable system designers to:
    (i) finetune for domain style and grounding, (ii) add adapters (e.g., LoRA) or tool-use interfaces, and (iii) experiment with retrieval-aware training curricula (e.g., instruction tuning with retrieved information).
    The trade-off is that model quality, safety alignment, and multilingual robustness may lag proprietary systems at equal compute budgets, and finetuning introduces data/compute requirements.
    \item \textbf{Retrieval-Native/Retrieval-Aware Generators.}
    These models explicitly integrate retrieval into model design and/or pretraining.
    Representative families include:
    (a) seq2seq RAG generators such as RAG~\citep{20neurips-rag} and FiD~\citep{21eacl-fid},
    (b) retrieval-augmented pretraining such as REALM~\citep{20icml-realm} and Atlas~\citep{23jmlr-atlas},
    (c) retrieval-enhanced decoder-only designs such as RETRO with chunked cross-attention over retrieved neighbors~\citep{22icml-retro},
    and (d) non-parametric memory augmentation such as kNN-LM~\citep{20iclr-knn-lm}.
    These approaches typically yield stronger information usage and easier knowledge updating by swapping the index, but require more complex training/inference pipelines and careful retriever--generator co-design.
\end{itemize}}

\edit{\subsection{How does Generator Design Impact RAG Performance}
Generator architecture shapes RAG performance along several axes:
\textbf{(i) Information sensitivity:} cross-attention or separate information encoding (FiD/RETRO) tends to use information more systematically than raw concatenation~\citep{21eacl-fid,22icml-retro}.
\textbf{(ii) Scaling with top-$k$:} models that can amortize information encoding and avoid full self-attention over all retrieved information tokens can scale to larger $k$, which is empirically beneficial in multi-document QA~\citep{21eacl-fid}.
\textbf{(iii) Long-context reliability:} long-context LLMs enable larger retrieved information budgets, but may still fail to access mid-context retrieved information, motivating ordering strategies and architectural fusion for robustness~\citep{23arxiv-lost-in-middle}.
\textbf{(iv) Grounding behavior:} retrieval-aware training (Atlas~\citep{23jmlr-atlas}, Self-RAG~\citep{23arxiv-self-rag}) improves factuality behavior and reduces hallucination relative to prompt-only RAG under similar retrieval settings.}

\edit{\subsection{Practical Guidelines for Selecting Generators}
Based on the above trade-offs, we provide selection guidelines according to different situations:
\begin{itemize}
    \item \textbf{Generator weights are inaccessible (API-only):}
    prioritize strong retriever/reranker, compact information formatting, and multi-stage prompting.
    Methods that tune retrieval while keeping the LLM frozen are suitable when you cannot change the generator~\citep{23arix-replug, chen2025beyond}.
    Long-context models help, but retrieved information should be positioned carefully due to long-context utilization issues~\citep{23arxiv-lost-in-middle}.
    \item \textbf{Need controllable grounding/citations or domain compliance:}
    choose open-weight generators and apply retrieval-aware finetuning (instruction tuning with retrieved information, contrastive supervision for citation, etc.), or use explicitly retrieval-aware generators (Atlas~\citep{23jmlr-atlas}, Self-RAG~\citep{23arxiv-self-rag}) when training resources allow.
    \item \textbf{Task needs many passages (multi-hop QA, long-form synthesis):}
    prefer architectures designed for multi-document fusion (FiD-style encoder-decoder or retrieval-native cross-attention) rather than prompt stuffing, because they scale better with top-$k$ and reduce attention dilution~\citep{21eacl-fid,22icml-retro}.
    \item \textbf{Latency/cost is critical:}
    select generators with inference-efficient attention or memory designs (e.g., GQA, sliding-window attention, KV compression, MoE) and keep retrieved context concise~\citep{23emnlp-gqa,guo2025deepseek}.
    \item \textbf{Knowledge updates are frequent:}
    retrieval-native/aware paradigms that separate parametric knowledge from an updatable index are preferable (RAG/REALM/Atlas/RETRO/kNN-style memory), since updating the corpus/index can be cheaper than continual finetuning~\citep{20neurips-rag,20icml-realm,23jmlr-atlas,22icml-retro,20iclr-knn-lm}.
\end{itemize}}

\edit{Overall, generator choice is not merely a matter of ``which LLM is strongest'', but of how the generator's architecture and training interface with retrieval, which directly determines information utilization, grounding quality, and system-level efficiency.}

%% file: 8-tasks.tex
\section{NLP Tasks}
\label{sec:task}

This section lists several classical tasks in the NLP domain, introduces advanced RAG techniques used to solve these tasks, and discusses the task-specifc challenges and failure modes 

\subsection{Language Modeling}

\edit{
\textbf{Task Characteristics.} 
Language modeling refers to the capability of modeling the probability distribution over sequences of tokens.
In practice, this capability is instantiated as the next-token prediction task: 
}
given such a sequence of tokens $x_1, \ldots, x_n$ called \textit{Prefix}, the language modeling task aims to model its probability via next-token prediction,
\begin{equation}
p(x_1, \ldots, x_n)=p(x_1)\cdot\prod^n_{i=2} p(x_i|x_1, \ldots, x_{i-1}),
\end{equation}
where the conditional probabilities $p(x_i|x_1, \ldots, x_{i-1})$ are modeled by a parameterized language model.

\edit{
\textbf{Technique Summary.} 
}
Recent works mainly leverage RAG further to improve language modeling capability in the pre-training stage. 
A branch of works~\citep{22icml-retro, 22neurips-encoder-decoder, 22iclr-mem-transformer, 23neurips-longmem} modifies the architecture of generators by adding a new cross-attention module in each transformer block for introducing retrieval knowledge. 
The intuition of those works is that given the similar \textit{Prefix}es and their next tokens (retrieving stage), the pre-trained model can calibrate the model's prediction using the cross-attention module to capture the pattern between the next token and prefix (model forwarding stage).
Zhong et al.~\citep{22emnlp-trime} propose to augment the language model with three types of retrieval memories/databases (local memory, long-term memory, and external memory) and optimize the next-token probability distribution with nearest neighbors retrieved from the memories/databases.
Another branch of works~\citep{20iclr-knn-lm, 23arxiv-knn-adapter, 23icml-knn-lm-why, 23arxiv-ralm, 20icml-realm} focuses on augmenting the inputs or outputs of generators with retrievals.
Guu et al.~\citep{20icml-realm} and Ram et al.~\citep{23arxiv-ralm} concatenate the retrieved information with inputs and feed the retrieval-augmented inputs into the generators.
Other works~\citep{20iclr-knn-lm, 23arxiv-knn-adapter, 23icml-knn-lm-why} fuse the logits of inputs as well as retrievals at the final output layer and generate the final probability distribution based on the interpolated results.
Those works believe that the concatenated/fused retrievals can provide useful context information on inputs/outputs to improve models' robustness during the pre-training stage.
Besides, Doostmohammadi et al.~\citep{23acl-retro-bm25} focus on pre-training models with a semantic retriever (BM25) and achieve a better language modeling performance.

\edit{
\textbf{Challenges and Failure Modes.}
A main challenge in retrieval-augmented language modeling is that the performance is highly sensitive to the retrieval quality.
This is because language modeling is evaluated by perplexity, a metric sensitive to every single token prediction. 
Even minor misalignment between the retrieved next token and the true next token would directly harm the performance, and the retrieval noise will be amplified rather than diluted.
Consequently, common failure modes can be:
(1) spurious copying of neighbors' next token that do not truly match the true next token, 
(2) noise amplification that shifts probability mass toward incorrect tokens and increases repetition or incoherence.
}

\subsection{Machine Translation}
\edit{
\textbf{Task Characteristics.} 
}
Machine translation (MT) leverages computational linguistics algorithms to translate text or speech from one language to another automatically. 
The goal of MT is to produce an accurate and fluent translation, preserving the meaning of the original text while adhering to the grammatical and stylistic norms of the target language. 
MT systems have evolved from rule-based machine translation (RBMT) to statistical machine translation (SMT) and, more recently, to neural machine translation (NMT).
In particular, NMT methods have significantly improved translation quality by leveraging deep learning techniques, which thus will be the focus of this section.

\edit{
\textbf{Technique Summary.} 
}
RAG techniques can further enhance MT by incorporating external knowledge into the translation process. 
The simplest way is to concatenate the similar translation examples into the inputs or fuse the logits of similar translation examples at the output layer.
For example, some works~\citep{22acl-reina, 23neurips-selfmem} retrieve similar translations according to the source text and concatenate corresponding target texts or pairs of source and target texts as examples into inputs. 
Other works~\citep{20acl-retrieve-edit-rerank, 21iclr-knn-mt, 21acl-ijcnlp-adaptive-knn-mt} feed the retrieved source text into the models and obtain the logits of the next target tokens, then aggregate all logits to generate the final predictions.
Moreover, Jiang et al.~\citep{22emnlp-robust-knn-mt} and Li et al.~\citep{23emnlp-source-context} use the logits of retrieved examples to calibrate the aggregated logits, improving the robustness of the generation.
Another branch of works~\citep{23acl-ink, 22emnlp-trime} injects external knowledge into the objective function during the training stage, refining the representation space with similar translations.
Besides, Cai et al.~\citep{21acl-monolingual-mem} encode similar translations and store them as the translation memory, then introduce the knowledge from memory with a cross-attention module.
Instead of improving the performance, a branch of work focuses on accelerating the generation efficiency on MT tasks, such as searching from a pre-built subset~\citep{22acl-fast-knn-mt, 23acl-subset-knn-mt} or a dynamic knowledge base~\citep{23iclr-sk-mt}, searching by chunks~\citep{22emnlp-chunk-knn-mt}.

\edit{
\textbf{Challenges and Failure Modes.}
In MT, retrieval augmentation must improve adequacy while preserving fluency and terminology consistency, which requires retrieving examples that are not only source-side similar but also reliable and stylistically compatible on the target side. 
However, mismatched examples can introduce subtle lexical/syntactic biases, and document-level consistency (e.g., domain terms and named entities) may not be guaranteed by sentence-level retrieval. 
Accordingly, failure modes often appear as:
(1) retrieval-biased terminology drift (incorrect or inconsistent term transfer), 
(2) over-copying retrieved target fragments leading to entity/number errors and reduced adequacy.
}

\subsection{Text Summarization}
\edit{
\textbf{Task Characteristics.} 
}
Text summarization is the process of condensing a larger text document into a shorter version, preserving key information and the overall message. 
This task can be broadly categorized into two types: extractive summarization, which involves selecting and compiling parts of the original text, and abstractive summarization, which entails rewriting the essence of the text in a new, concise form. 
The goal is to produce a coherent and fluent summary that encapsulates the most critical information from the source material.

\edit{
\textbf{Technique Summary.} 
}
RAG techniques can significantly enhance text summarization tasks by leveraging external knowledge and similar documents to inform the summarization process. 
\citep{22acl-reina, 22acl-retrieval-bio, 23acl-udr, 23neurips-selfmem} simply concatenates the retrieved similar summaries into inputs to generate summarizations.
Instead of concatenating texts, other works fuse features at the intermediate layers by cross-attention~\citep{23neurips-unlimiformer}, or at the output layers by logits ensemble~\citep{20acl-retrieve-edit-rerank}.
Besides, Jiang et al.~\citep{23emnlp-flare} argue that retrieving for every generation may not always be the best choice and propose to retrieve external knowledge during the generation process adaptively.

\edit{
\textbf{Challenges and Failure Modes.}
For summarization, the central challenge is balancing compression with grounding: retrieval can supply helpful external information, but it also tends to introduce redundancy and tangential details that compete with the source document for attention under a limited context budget. 
Furthermore, summarization often requires abstraction and synthesis rather than copying, so naïvely injecting retrievals can bias the model toward extractive behavior. 
As a result, typical failure modes include 
(1) content contamination, where irrelevant retrieved facts leak into the summary and reduce faithfulness, 
(2) retrieval accuracy, where the summarization task often requires document-level retrievals which may occur semantic drift issues, 
and (iii) copy bias that harms abstraction and can even import contradictions from noisy retrievals.
}

\subsection{Question Answering}
\edit{
\textbf{Task Characteristics.} 
}
Question Answering (QA) is a fundamental task in NLP that involves building systems capable of automatically answering human questions in natural language. 
QA systems can be broadly classified into two categories: open-domain, where the system answers questions about virtually anything, and closed-domain, focusing on a specific area of knowledge. 
Due to the page limits, this paper only discusses the works of open-domain QA systems.

\edit{
\textbf{Technique Summary.} 
}
RAG techniques combine information retrieval with model-based generation, which is highly suitable for QA systems.
In particular, open-domain QA systems usually first require searching for knowledge from the Internet or large-scale databases, then generate the corresponding answers according to the retrieved information.
Naturally, given similar questions and corresponding answers as demonstrations which are concatenated into inputs~\citep{22acl-reina, 23acl-udr, 23arxiv-raven}, generators in RAG can learn the pattern between questions and answers and infer what answers should be.
For some specific QA tasks where a set of reference documents is given, retrievers in RAG would retrieve the relevant documents for concatenation, and then generators in RAG would read the context then generate the final answers via the self-attention mechanism~\citep{20icml-realm, 23acl-ur-qa, 23arxiv-ralm, 23arxiv-self-rag}, which is similar to solving a reading comprehension problem.
Besides, Fabbri et al.~\citep{20acl-templated-retrieval} focus on designing effective templates for re-organizing the concatenated contexts.
Baek et al.~\citep{23arxiv-kaping} leverage the knowledge graph to retrieve the related facts for the input questions, then feed their concatenation and inputs into the generators.
Instead of directly concatenating texts, another branch of works focuses on joining the retrieval embeddings with input embeddings for the encoder-decoder models~\citep{21eacl-fid, 21neurips-emdr2, 23icml-lumen, 23jmlr-atlas}.

Some works incorporate the external knowledge in the hidden states or the final logits of generators.
For the fusion in the hidden states, the key is what kind of knowledge representation would be injected, such as entities~\citep{20emnlp-eae, 22iclr-tome}, chunks~\citep{22icml-retro, 23arxiv-retrio++}, documents~\citep{23arxiv-pluglm}. 
For the fusion in the logits, most works combine the logits of retrievals and inputs by ensemble techniques~\citep{23arix-replug, 20icml-realm, 20neurips-rag, 23acl-demrag}.

Instead of designing different knowledge fusions for QA systems, existing works also improve QA systems with RAG from other aspects.
Some works~\citep{19acl-coupling-retrieval, 22acl-rgf, 22neurips-recross} use retrieved question-answering pairs as extra training data.
Some works optimize the retriever module, e.g., improving the keys' representation when building the retriever database~\citep{23acl-dist-over-vocab}, replacing the indexing with a pre-trained ranking model~\citep{23acl-aar}, or enabling retrieving phrases with two queries~\citep{23acl-npm}.
Other works focus on accelerating the generation efficiency of RAG.
Jong et al.~\citep{22arxiv-fido} propose the layer-sparse cross-attention to speed up the decoding.
Some works~\citep{23arxiv-self-rag, 23emnlp-flare, 23emnlp-skr} observe that the retrievals may not always provide useful information during the generation process and learn to determine when to retrieve.
Moreover, Sun et al.~\citep{24iclr-tog} combine the RAG with agents to iteratively reason the final results.

\edit{
\textbf{Challenges and Failure Modes.}
Open-domain QA is highly sensitive to retrieval quality because those tasks require the retrieved passages to contain the exact answer span.
In addition, QA systems are frequently expected to provide verifiable grounding (e.g., attribution/citations) and to abstain when retrieved information is insufficient, which further raises the bar for information utilization.
Accordingly, failure modes commonly manifest as: 
(1) missing-information hallucination, where the model answers from parametric memory when the retriever fails to obtain the passage that contains answers, 
(2) cherry-picking under contradiction, where the model selectively uses one snippet while ignoring conflicting retrieved information.
}

\subsection{Information Extraction}
\edit{
\textbf{Task Characteristics.} 
}
Information Extraction (IE) is a critical task in NLP to automatically extract structured information from unstructured and semi-structured text sources. 
This task encompasses several sub-tasks, including Named Entity Recognition (NER), Entity Linking (EL), Coreference Resolution (CR), Relation Extraction (RE), etc. 
The goal is to identify and classify key elements from text and understand the relationships between them, thereby converting textual data into a structured format amenable to analysis and interpretation.

\edit{
\textbf{Technique Summary.} 
}
With RAG techniques, addressing IE tasks can be significantly improved in terms of not only performance but also interpretability.
In NER tasks, Wang et al.~\citep{21acl-cl-kl} first retrieve similar sentences and then concatenate the ranked retrievals for better semantic representations.
Ren et al.~\citep{23acl-retrieve-sample} show that naive RAG may not address Event Argument Extraction (EAE) tasks. 
Thus, they adopt a sampling-based method to guarantee the same distribution of event labels between retrievals and inputs then concatenate retrieval texts into inputs for better performance in EAE tasks.
Table augmentation is also a challenging task, which requires extracting information from tables.
Glass et al.~\citep{23acl-rata} propose to leverage a reconstruction-based method with a trained retrieval-based model to address table augmentation.

\edit{
\textbf{Challenges and Failure Modes.}
The challenge in applying RAG to information extraction lies in how to effectively integrate structured external knowledge, such as knowledge graphs for entities and relations or table data. 
Simple concatenation fails to fully convey the relational and type information inherent in such structured data, necessitating more sophisticated fusion mechanisms.
Besides, useful retrievals must be not only semantically similar but also label-compatible (e.g., entity types, relation schemas, event templates), otherwise the retrieved patterns can mislead span boundaries and role assignments. 
Consequently, failure modes often appear as:
(1) information injection failure, where those structured information cannot be effectively injected into the model for accurately extraction, 
(2) schema/label drift, where retrievals reflecting different conventions degrade extraction consistency.
}

\subsection{Text Classification}
\edit{
\textbf{Task Characteristics.} 
}
\edit{Text classification tasks are common in NLP applications~\citep{25ftir-classification}.}
Sentiment analysis, a prominent text classification task in NLP, entails identifying and categorizing the emotional tone conveyed in a text.
For example, given a sentence of ``I love to watch movies'', the analysis models should determine whether it has a positive attitude or a negative attitude.
The attitude in sentiment analysis can range from positive to negative or can be neutral, nuanced, and even mixed. 
The sentiment analysis task is crucial for understanding consumer feedback, monitoring brand reputation, and gaining insights into public opinion on various issues.

\edit{
\textbf{Technique Summary.} 
}
RAG techniques can significantly enhance sentiment analysis with different external knowledge fusion strategies.
Li et al.~\citep{23acl-udr} concatenate the retrieved options and corresponding prompt-based labels with input options.
Other works~\citep{22neurips-retroprompt, 23acl-pgra} concatenate the retrieval embeddings with input embeddings before feeding them into the decoder.
Some works fuse the retrieval features into the hidden states of generators via cross-attention~\citep{23arxiv-pluglm, 23neurips-longmem} or ranking-based addition~\citep{24iclr-refusion}.
Besides, other works focus on fusing the logits of retrievals with the output logit using ensemble techniques~\citep{23acl-reaugkd, 23emnlp-eml}.
Except for knowledge fusions, Min et al.~\citep{23acl-npm} enable locating knowledge in phrases more accurately via two queries.

\edit{\textbf{Challenges and Failure Modes.}
For text classification, retrieval noise can directly mislead the classification, as even semantically similar documents may carry different labels. 
Moreover, the problem is exacerbated under domain shift, where retrieval may reinforce dataset artifacts rather than true label semantics. 
Accordingly, failure modes typically manifest as:
(1) nearest-neighbor confusion, where retrieved instances are similar but differently labeled and systematically bias predictions, 
(2) demonstration-induced bias/instability, where predictions fluctuate with changes in retrieved examples or their ordering.}

\subsection{Dialogue Systems}
\edit{
\textbf{Task Characteristics.} 
}
Dialogue systems, also known as conversational agents or chatbots, are designed to simulate conversation with human users, either in text or speech form. These systems can be categorized into two main types: task-oriented systems~\citep{ra_pdg}, which assist users in completing specific tasks such as booking tickets or ordering food, and open-domain systems, which aim to carry on a general conversation on a wide range of topics~\citep{ra_hc}. The core challenge in developing effective dialogue systems lies in understanding user intent, maintaining context, and generating coherent, relevant responses.

\edit{
\textbf{Technique Summary.} 
}
Existing works improve the dialogue system with RAG mostly via the query-based fusions.
Some works~\citep{21acl-mamd, 23acl-refpydst, 23neurips-selfmem} concatenate the retrieved history conversations with current inputs.
Other works~\citep{21tacl-kif, 23acl-recap, 23neurips-selfmem} first leverage an encoder to encode the history responses, then feed the concatenated embeddings into a decoder to generate new responses.

\edit{
\textbf{Challenges and Failure Modes.}
Dialogue systems operate under multi-turn context accumulation, making retrieval augmentation challenging in selecting what to retrieve (history, profile, external knowledge) without overwhelming the context window or harming conversational flow, while also respecting privacy and latency requirements. 
In this setting, failure modes commonly manifest as: 
(1) stale or incorrect memory grounding, which causes inconsistencies across turns, 
(2) evolution of user intent, where user intent may change dynamically during the conversation.
}

%% file: 7-benchmarking.tex
\section{RAG Evaluation and Benchmark}
\label{sec:eval}

\edit{The evaluation of RAG systems is inherently more complex than evaluating standalone retrievers or language models. Performance hinges on the synergistic interaction between two core components: the retriever's ability to fetch relevant, context-supporting information, and the generator's ability to faithfully and accurately synthesize this knowledge into a coherent response. Consequently, effective evaluation must disentangle these contributions while also assessing the integrated system's end-to-end efficacy. This section provides a comprehensive taxonomy of RAG-specific metrics, analyzes fundamental evaluation challenges, compares prevailing methodologies, and critically examines the limitations of existing benchmarks to illuminate paths for future work.}

\begin{table}[t]
\centering
\caption{\textbf{Taxonomy of RAG Evaluation Metrics and Dimensions.} The taxonomy organizes metrics by their primary evaluation focus, moving from classic retrieval to RAG-specific generation and robustness, culminating in system-level operational concerns. Note: Answer Faithfulness is distinct from Answer Correctness; an answer can be faithful to an incorrect retrieved context.}
\begin{tabular}{p{0.18\linewidth} p{0.17\linewidth} p{0.3\linewidth} p{0.28\linewidth}}
\toprule
\textbf{Evaluation Focus} & \textbf{Dimension} & \textbf{Metrics} & \textbf{Representative Works} \\
\midrule
\textbf{Retrieval Quality} & Context Relevance & Precision@k, Recall@k, Hit Rate, Mean Reciprocal Rank (MRR), Normalized Discounted Cumulative Gain (NDCG) & TREC, MS MARCO, BEIR~\citep{24nips-thakur-beir}, Domain-specific adaptations~\citep{24arxiv_legal_ragbench} \\
\midrule
\textbf{Generation Performance} & Answer Faithfulness (Groundedness) & Answer Attribution / Faithfulness Score, 
\newline Citation Precision/Recall & RAGAS~\citep{24eacl-es-ragas}, ARES~\citep{24naacl_ares}, FActScore~\citep{23emnlp-min-factscore}, TRUVA~\citep{25arxiv-truth} \\
\cmidrule(lr){2-4}
& Answer Relevance & Answer Similarity (BERTScore, ROUGE), LLM-as-a-Judge relevance scoring & RAGAS~\citep{24eacl-es-ragas}, AlpacaEval~\citep{24arxiv-alpacaeval} \\
\cmidrule(lr){2-4}
& Answer Correctness & Exact Match (EM), F1 Score, Task-Specific Accuracy& OpenQA~\citep{23arxiv-rag-openqa}, Domain-specific (e.g.,~\citep{24acl_medical_ragbench, 24nips_ragbench}) \\
\midrule
\textbf{End-to-End Robustness} & Noise Robustness, Negative Rejection, Information Integration, Counterfactual Robustness & Accuracy, Rejection Rate, Error Detection Rate, Error Correction Rate & RGB~\citep{24aaai_ragbench} \\
\midrule
\textbf{System-Level Efficiency} & Operational Performance & Latency (p95, p99, TTFT), Cost (USD per query), Throughput, Memory Footprint & Refusion~\citep{24iclr-refusion}, ACIN~\citep{25nejm-acin}, HedraRAG~\citep{25arxiv-hedrarag} \\
\bottomrule
\end{tabular}
\label{tab:rag-taxonomy}
\end{table}

\edit{\subsection{A Taxonomy of RAG Evaluation}
Table~\ref{tab:rag-taxonomy} provides a consolidated overview of RAG evaluation metrics, organized by evaluation focus.
We categories them into four evaluation focus, including retrieval quality, generation performance, end-to-end robustness, and system-level efficiency.
}

\edit{\textbf{Retrieval Quality} assesses the quality of the retrieved context. \textit{Context Relevance}, the primary dimension here, encompasses traditional information retrieval metrics—such as Precision@k, Recall@k, and NDCG—that measure the topical alignment between retrieved passages and the query.}

\edit{\textbf{Generation Performance} focuses on the final output across three distinct dimensions. \textit{Answer Faithfulness} (or Groundedness) measures whether the generated answer is entirely supported by the provided context, serving as a critical guard against hallucinations~\citep{24tacl_retriever_eval, 24eacl-es-ragas}. \textit{Answer Relevance} assesses whether the response addresses the user's query and remains on-topic, while \textit{Answer Correctness} measures factual accuracy against a ground truth answer~\citep{25arxiv_ragbench}. These dimensions are conceptually distinct: an answer can be faithful to retrieved information yet factually incorrect if the information itself is wrong, underscoring the need for comprehensive evaluation across all three.}

\edit{\textbf{End-to-End Robustness} extends beyond basic quality metrics to evaluate system behavior under challenging conditions. Following frameworks like RAGBench~\citep{24aaai_ragbench}, this dimension encompasses several distinct ``abilities'': \textit{Noise Robustness} (handling irrelevant or distracting context), \textit{Negative Rejection} (refusing to answer when retrieved information is insufficient), \textit{Information Integration} (synthesizing multiple retrieved information pieces), and \textit{Counterfactual Robustness} (resisting misleading information). Recent benchmarks such as RaLLe~\citep{24emnlp-hoshi-ralle} provide targeted datasets to stress-test these capabilities.}

\edit{\textbf{System-Level Efficiency} captures operational considerations that determine deployability. Beyond answer quality, practical RAG systems must satisfy constraints on \textit{latency} (particularly tail latencies such as p95 and p99, as well as time-to-first-token), \textit{cost} (e.g., USD per query), \textit{throughput} (queries per second under fixed service-level objectives), and \textit{memory footprint} (including index size and cache overhead). These metrics enable bottleneck attribution and guide architectural choices. For instance, ACIN~\citep{25nejm-acin} reports per-query costs, ReFusion~\citep{24iclr-refusion} demonstrates latency reductions through architectural fusion, and HedraRAG~\citep{25arxiv-hedrarag} emphasizes cross-stage co-optimization to improve end-to-end latency and throughput predictability.}

\edit{\subsection{Key Evaluation Challenges}
The interdependency of retrieval and generation creates unique evaluation pitfalls.}
\edit{One key challenge is the trade-off between context faithfulness and context relevance~\citep{23cs-hallucination-survey}. The faithfulness–relevance trade-off highlights a fundamental tension in text generation. An answer that is strictly faithful to the provided context may still be of low quality if the context is only marginally relevant to the question. Conversely, an answer that is highly relevant to the user’s intent may require reasoning beyond the given context, thereby sacrificing faithfulness. The optimal balance between faithfulness and relevance is inherently application-dependent. For instance, legal or medical question answering prioritizes strict faithfulness to source documents~\citep{23chi-eric-medicalqa, 20emnlp-2020-legal}, whereas creative writing or brainstorming tasks may tolerate, or even benefit from, reduced faithfulness in favor of higher relevance and usefulness.}

\edit{Another significant issue is the difficulty of determining whether an error originates from the retriever or the generator. Specifically, an incorrect answer may result either from the retriever’s failure to surface relevant or sufficient retrieved information, or from the generator’s failure to faithfully interpret and utilize otherwise correct retrieved information~\citep{20neurips-rag, 24nips-ru-ragchecker}. Disentangling these two sources of error is non-trivial, as both components are tightly coupled during inference and are typically evaluated only through the final generated output~\citep{25arxiv_ragbench, 24nips-ru-ragchecker}. Accurate attribution is nevertheless crucial for effective system debugging, evaluation, and model improvement. Without such attribution, designing targeted interventions—such as improving retrieval quality, enhancing grounding constraints, or refining decoding strategies—becomes challenging. However, reliable attribution often requires fine-grained annotations, such as information-level relevance judgments or claim-level faithfulness labels, which are expensive and time-consuming to obtain, particularly in domain-specific settings like legal or medical question answering. As a result, many existing benchmarks and evaluations conflate retrieval and generation errors, obscuring the true source of system failures and limiting the interpretability of empirical results.~\citep{23emnlp-autoeval, 24acl-li-attributebench}}

\edit{A third challenge concerns the correlation of automatic metrics with human judgment. In open-ended RAG and other long-form generation settings, traditional lexical-overlap metrics such as BLEU and ROUGE are often unreliable proxies for human-perceived answer quality: meta-evaluation studies report weak or inconsistent correlations with human ratings on key dimensions like coherence and relevance. Overlap-based scoring can also be misleading for certain question answering or machine reading comprehension answer types (e.g., yes/no or entity-list questions)~\citep{21tacl-summeval, 20acl-bleurt}. Moreover, recent RAG-specific analyses note that n-gram–based metrics may work for short answers but fail to capture fine-grained quality differences in longer, multi-claim responses, motivating more diagnostic evaluation protocols~\citep{24nips-ru-ragchecker}. These limitations have contributed to a shift toward model-based, reference-free evaluation, particularly the LLM-as-a-judge paradigm, which uses strong LLMs to grade responses under explicit rubrics and has demonstrated substantially better alignment with human preferences on open-ended tasks~\citep{23emnlp-geval, 23nips-llmjudge}.}

\edit{Evaluating unknown or refusal responses is also challenging in open-ended RAG systems. A reliable model should abstain (e.g., answer ``I don't know'' or refuse) when the retrieved information is missing, contradictory, or irrelevant to the query, rather than hallucinating. However, measuring this capability requires carefully constructed negative instances—such as adversarially unanswerable questions or deliberately flawed contexts—beyond standard answerable QA data \citep{18acl-squad2,21emnlp-know,25arxiv-refusebench}. Moreover, the metric must avoid rewarding trivial over-refusal (e.g., refusing everything): selective prediction formulations make this trade-off explicit by jointly evaluating accuracy (risk) and the fraction o0f questions answered (coverage), for example using risk--coverage curves or coverage at a fixed risk level \citep{21acl-abstention}.}

\edit{\subsection{Comparison of Evaluation Methodologies}
RAG systems can be evaluated using a spectrum of methodologies, each trading off validity, cost, and scalability. In practice, robust assessment often combines multiple methods rather than relying on a single score.}

\edit{\textbf{Human Evaluation:} The traditional gold standard, where trained annotators rate dimensions such as fluency, factual correctness, relevance, and faithfulness/grounding to the retrieved information. When carefully designed with clear rubrics and quality control, it offers high validity and diagnostic value. However, it is expensive, slow, and difficult to scale, making it impractical for rapid iteration or large-scale benchmarking.}

\edit{\textbf{Automatic Metric-Based Evaluation:} Relies on predefined, programmatic metrics (e.g., EM, token-level F1, BERTScore) for fast and reproducible scoring~\citep{23arxiv-rag-openqa}. This approach is scalable and inexpensive, but it can under-measure semantic adequacy and holistic answer quality in open-ended generations, and it often fails to directly capture information-grounding or multi-claim factual consistency.}

\edit{\textbf{LLM-as-a-Judge:} An increasingly prevalent paradigm in which a strong LLM (e.g., GPT-4) is prompted to score, critique, or rank system outputs against explicit criteria such as faithfulness and relevance. It can serve as a scalable proxy for human judgments, particularly for nuanced, rubric-based criteria that are hard to encode as simple metrics. Nevertheless, it introduces additional inference cost and may inherit biases from the judge model (e.g., reliance on its parametric knowledge, sensitivity to prompt phrasing, or limited objectivity). Frameworks such as ARES~\citep{24naacl_ares} aim to systematize and standardize this evaluation workflow.}

\edit{\textbf{End-User/Task-Oriented Evaluation:} The most ecologically valid form of evaluation, measuring performance within a real workflow (e.g., reduced time-to-completion, improved decision quality, or higher user satisfaction)~\citep{22chi-llm-code-eval}. While highly meaningful for deployment, it is context-dependent, costly to run, and difficult to generalize across domains or tasks.}

\edit{\subsection{Benchmark Limitations and Critical Gaps}
Despite rapid progress in RAG evaluation, several structural limitations and open gaps remain that hinder reliable system comparison and principled iteration~\citep{24arxiv-rag-benchmark}.}

\edit{\textbf{Over-Reliance on Static, Wikipedia-Centric Knowledge.}
A large fraction of widely-used QA/RAG benchmarks are ultimately grounded in snapshots of Wikipedia (e.g., Natural Questions~\citep{19tacl-nq}, HotpotQA~\citep{yang2018hotpotqa}), where questions are annotated against specific Wikipedia pages or paragraphs. 
While these datasets are valuable for controlled benchmarking, they weakly reflect primary production settings where RAG is deployed over dynamic, proprietary, or domain-specific knowledge bases (e.g., continuously updated technical documentation, internal corporate wikis, ticketing systems, or policy repositories). 
The domain-focused benchmarks and corpora such as TechQA~\citep{2020acl-techqa},  MIRAGE for medical RAG \citep{24acl_medical_ragbench}, and RAGBench with industry corpora such as user manuals \citep{25arxiv_ragbench} broaden coverage beyond Wikipedia, but overall benchmark diversity is still limited and often fails to capture the full variability of enterprise data and update dynamics.}

\edit{\textbf{Lack of Fine-Grained, Diagnostic Benchmarks.}
Many benchmarks collapse performance into a single aggregate score, which is insufficient for debugging modular RAG pipelines. 
A key missing capability is diagnostic evaluation that supports error attribution: distinguishing whether a failure is caused by retrieval (e.g., missing the crucial passage), comprehension (e.g., misreading a number/date), or reasoning (e.g., failing to integrate multiple retrieved information snippets). 
Recent efforts move toward fine-grained diagnosis---for example, RAGChecker introduces module-aware diagnostic metrics \citep{24nips-ru-ragchecker}. 
However, such diagnostic supervision typically requires more granular annotation (e.g., retrieved information mapping, claim-level grounding), which is expensive and difficult to scale.}

\edit{\textbf{Under-exploration of Efficiency and Cost.}
Evaluation still overwhelmingly prioritizes answer quality while under-reporting operational metrics that determine deployability: end-to-end latency (including p95/p99), time-to-first-token (TTFT), throughput, memory footprint, and dollar cost per query. 
Systems studies show that RAG introduces non-trivial latency--throughput--memory trade-offs and can substantially increase TTFT and storage demands, making quality-only comparisons incomplete \citep{shen2024towards,jiang2025rago}. 
Consequently, a ``better'' RAG method under offline quality metrics may be impractical under real serving constraints.}

\edit{\textbf{Insufficient Focus on Multi-Modal and Complex Retrieval.}
As RAG expands beyond plain text to PDFs with tables/figures, images, and structured backends (e.g., SQL databases), evaluation frameworks lag behind and lack consistent, standardized protocols across modalities. 
Emerging benchmarks highlight this gap, such as MRAG-Bench for vision-centric retrieval-augmented multimodal models \citep{hu2024mrag} and the multimodal RAG Benchmark built from PDF pages ~\citep{peng2025unidoc}. 
For table-heavy and heterogeneous settings, recent benchmarks and tasks (e.g., TARGET for table retrieval \citep{ji2025target}, and TableRAG for retrieval plus SQL-style reasoning over heterogeneous documents \citep{yu2025tablerag}) represent important progress, but evaluation remains fragmented and modality-specific rather than unified.}

\edit{\textbf{Temporal Dynamics and Knowledge Freshness.}
Few benchmarks explicitly test temporal reasoning and the ability to prioritize up-to-date information, despite this being crucial for real deployments (e.g., news, finance, policy compliance). 
Time-sensitive QA benchmarks and protocols such as the Time-Sensitive QA dataset \citep{chen2021dataset}, FreshQA~\citep{23arxiv-refreshllms}, and contamination-resistant UnSeenTimeQA \citep{uddin2025unseentimeqa}) show that models often struggle when facts evolve or questions require temporal debunking.
On the retrieval side, recent benchmarks such as TEMPO explicitly combine temporal reasoning with reasoning-intensive retrieval and introduce temporal-specific retrieval metrics \citep{abdallah2026tempo}, but such evaluations are still not mainstream in end-to-end RAG benchmarking.}

\edit{Overall, RAG evaluation is shifting from reusing classical information retrieval and question answering metrics toward multi-dimensional, module-aware frameworks that emphasize grounding/faithfulness, robustness, and diagnostic usefulness \citep{24arxiv-rag-benchmark,24nips-ru-ragchecker,25arxiv_ragbench}. 
Future evaluation suites should further close the gaps in (i) diagnostic error attribution, (ii) efficiency--cost reporting under realistic serving constraints, (iii) multimodal and structured retrieval, and (iv) temporal freshness. 
A robust evaluation suite is not only an assessment instrument but also a development roadmap for building RAG systems that are reliable, efficient, and trustworthy.}

%% file: 6-training.tex
\section{RAG Training and KB Update}
\label{sec:training}

This section introduces RAG training, which can be categorized into two main classes: \textbf{RAG without knowledge base update} and \textbf{RAG with knowledge base update}.
The former refers to the case where only trainable parameters in each module of RAG would be updated, and the knowledge in the KB would remain the same during the training stage.
The latter refers to the case where the knowledge in the knowledge base would be updated, then each module's parameters in RAG would be updated in a similar way as the former case.

\subsection{RAG without Knowledge Base Update}
\begin{figure*}[t]
\centering
\includegraphics[width=1.0\linewidth]{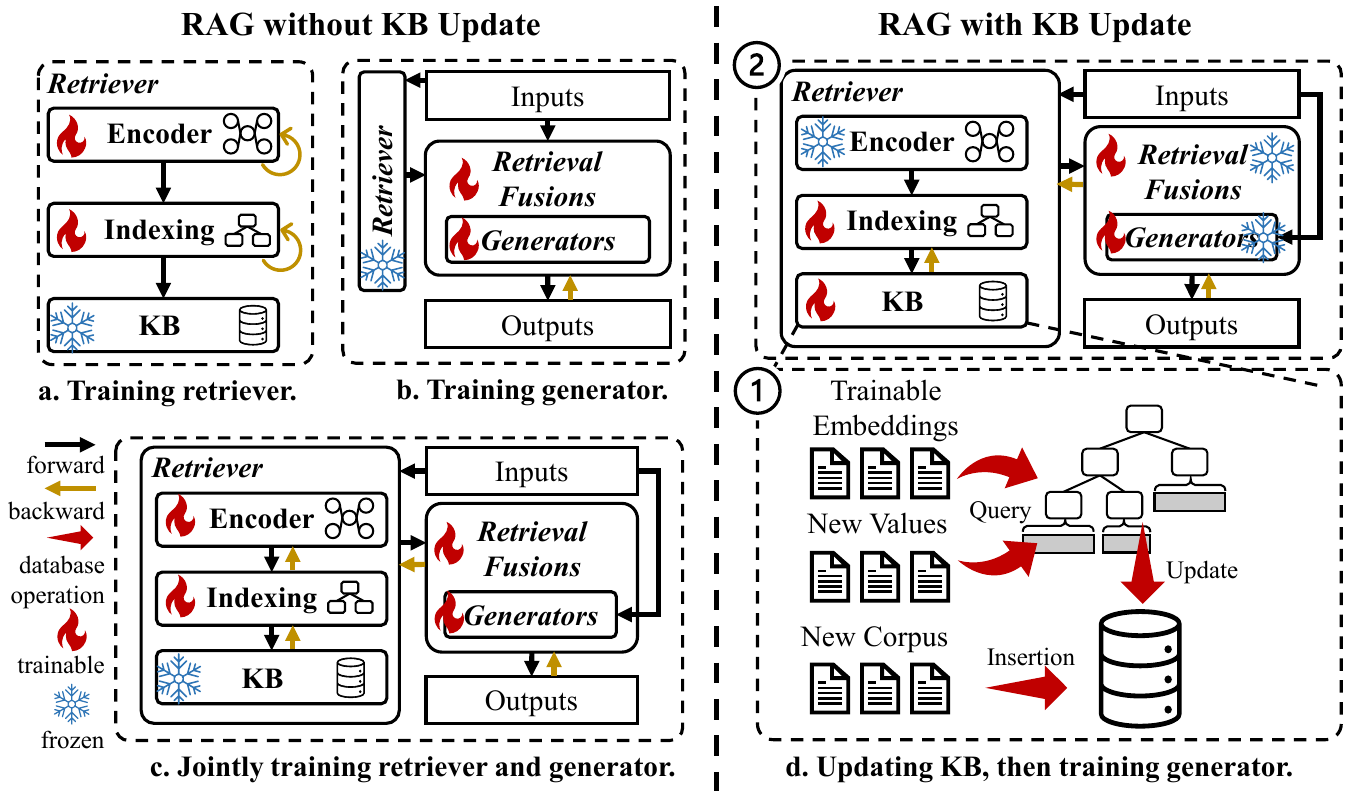}
\caption{Different RAG training strategies with/without knowledge base update.}
\label{fig:update}
\end{figure*}

The goal of training RAG without a knowledge base update is to update the knowledge stored in the short-term memory of generators based on the existing knowledge base.
As shown in Figure~\ref{fig:update} (a)-(c), there are three training cases, i.e., training the retriever, training the generator, and jointly training the retriever and generator.

\subsubsection{Training Retriever.}
Considering the case of no knowledge base update, training the retriever generally refers to training the retriever encoder and rebuilding the indexing.
Since sparse encodings rely on statistical methods without parameters, training the encoder pertains only to dense encoding methods.
Different training methods may have different goals, such as improving the semantic representations, accelerating the encoding process, or learning the domain-specific representations.
The first two goals are often achieved by replacing the original encoder with a more powerful or tiny encoder, such as DistilBERT~\citep{19arxiv-distilbert}, or TinyBERT~\citep{20emnlp-tinybert}.
The last requires training the original encoder on the domain-specific corpus.
REPLUG~\citep{23arix-replug} updated the retriever by minimizing KL divergence between retrieval and LM scores, and asynchronously refreshing the knowledge base index.
After training the retriever encoder, the vector database's embeddings that serve as keys will also change.
Thus, all indexes should be rebuilt with new embeddings.
Besides, if the encoder remains unchanged, the indexing can be updated using new ANN searching algorithms or re-tuning the hyperparameters.
After the retriever is trained, it can be directly incorporated into the RAG without updating the generator.

\subsubsection{Training Generator.}
Training the generator involves updating its parameters or those in the retrieval fusion modules.
Since the generator is generally an LLM, training the LLM is a resource- and time-consuming process.
Fortunately, several parameter-efficient fine-tuning techniques, such as LoRA~\citep{22iclr-lora}, are proposed to address the fine-tuning problem of LLMs.
Although the parameters in the retrieval fusion modules are less than those in the generator, only fine-tuning those parameters may encounter some training problems, such as low convergence and overfitting.
Jointly tuning the parameters in the generator and the retrieval fusion modules is a better way to train the generator and the retrieval fusion modules if there are sufficient and powerful resources.
\subsubsection{Jointly Training the Retriever and Generator.} 
Apart from independently training the retriever and the generator, jointly training the retriever and the generator can be another good choice for better performance on downstream tasks.
The primary challenges involve computational complexity from large-scale retrieval and marginalization over latent documents, alongside training instability due to potential retrieval collapse and interdependent retriever-generator feedback loops.
Additionally, sparse gradients from discrete retrieval steps and approximation errors from fixed top-$k$ document selection limit effective end-to-end optimization. 
Typically, complex indexes, such as FAISS~\citep{24arxiv-faiss}, are not a suitable choice during the fine-tuning stage.

Existing works generally leverage the complex indexes to pre-select a small subset of nearest neighbors as candidates, then choose the final top-$k$ nearest neighbors by performing the matrix-multiplication operations.
REALM~\citep{20icml-realm} and RAG~\citep{20neurips-rag} both unify retriever and language modeling by treating retrieved documents as latent variables and optimizing end-to-end via gradient descent. 
They both used Maximum Inner Product Search (MIPS).
RAG~\citep{20neurips-rag} only updated the encoder in the retriever, but REALM~\citep{20icml-realm} also asynchronous refreshed the indexing. 
Atlas~\citep{23jmlr-atlas} jointly pre-trained the retriever and generator, using different loss functions (such as attention distillation and perplexity distillation) to optimize the retriever. 
Besides, it also asynchronous refreshed the indexing. 
Experimental results in these works demonstrate that joint training is an end-to-end optimization that can lead to better coordination between the retriever and the generator and improve the contextual understanding of the generator. 

\subsection{RAG with Knowledge Base Update}
As shown in Figure~\ref{fig:update} (d), the scenario involves two stages: updating the knowledge base, then training the retriever and the generator.
There are three cases for updating the knowledge base, i.e., updating with trainable embeddings, updating with new values, and updating with a new corpus.
In the first case, values generally are trainable embeddings and are simultaneously/asynchronously updated with parameters in the RAG~\citep{22neurips-retroprompt}.
The last two cases usually refer to updating the knowledge base with up-to-date information.
Taking the question-answer corpus as an example, updating with new values refers to updating the answer to existing questions, while updating with a new corpus refers to adding new question-answer pairs.
To update the value of existing keys requires first querying the existing key-value pairs and then performing in-place updates.
For a new corpus, the knowledge base first needs to perform insertion operations, then rebuild or update the indexes as new keys are added.
After updating the knowledge base, training the retriever and the generator is similar to RAG without knowledge base updates.
However, this training step is not always necessary, thanks to the in-context learning capability of LLMs.

%% file: 9-applications.tex
\section{Applications}
\label{sec:app}

\edit{\subsection{RAG in Industry: An Agent-Centric View}}
LLM-based autonomous agents are intelligent software systems that leverage the power of LLMs to perform tasks without continuous human intervention~\citep{li2024personalllmagentsinsights, xi2023risepotentiallargelanguage, Wang2024}.
These agents use LLMs as a brain or controller~\citep{huang2024understandingplanningllmagents}, and extend their abilities through multimodal perception~\citep{xie2024largemultimodalagentssurvey}, tool utilization~\citep{NEURIPS2023_d842425e}, and external memory~\citep{packer2024memgptllmsoperatingsystems}.
\edit{In industrial deployments, RAG is widely adopted as the grounding layer for agentic systems (copilots/assistants), because it enables agents to (i) consult proprietary knowledge with provenance and (ii) update knowledge without retraining the model~\citep{20neurips-rag, azure_rag_overview}.
Moreover, recent enterprise RAG practices increasingly emphasize governed retrieval rather than unrestricted open-web browsing~\citep{m365_copilot_retrieval_api}.}

\textbf{Using RAG to Retrieve from External Memory.}
LLM-based agents can utilize RAG to access and retrieve information from their own external memory~\citep{hatalis2023memory, zhang2024surveymemorymechanismlarge, mei2024aiosllmagentoperating}.
This external memory serves as a knowledge base that the agent can draw upon to enhance its understanding and decision-making.
When faced with a query or a task, the agent retrieves relevant memory items and integrates them into the generation process of the LLM.
This allows the agent to produce responses or solutions informed by a wider range of knowledge, leading to more accurate and contextually relevant outcomes.

\textbf{Using Tools and RAG for Up-to-Date Information.}
\edit{In addition to retrieving information from its own memory, an agent can use tools to acquire fresh information, which is useful for tasks requiring up-to-date knowledge (e.g., market and policy updates)~\citep{NEURIPS2023_d842425e}.
In enterprise settings, such retrieval is often implemented via governed interfaces (e.g., enterprise search/connectors) to preserve permissions and compliance.
For example, the Microsoft 365 Copilot Retrieval API powers applications that access content from SharePoint, OneDrive, and Copilot connectors while maintaining secure access controls.
For complex queries in copilot apps, agentic retrieval further decomposes a question into multiple focused subqueries and aggregates results for downstream answer synthesis~\citep{m365_copilot_retrieval_api}.
Overall, RAG augments agent capabilities by enabling access to broader and fresher information from memory and external sources, improving decision-making and reliability.}

\edit{\subsection{Frameworks and Libraries for RAG Application Development}
The practical implementation of RAG has been greatly accelerated by the emergence of specialized software libraries and orchestration frameworks. While often colloquially referred to as platforms, these tools are more accurately described as development frameworks that provide the necessary primitives for building, managing, and optimizing RAG pipelines.}

\edit{The foundational libraries in this space include LangChain~\citep{langchain} and LlamaIndex~\citep{22github-llamaindex}, which pioneered the modular composition of retrieval, augmentation, and generation components. Since their introduction, the ecosystem has diversified significantly to accommodate different architectural preferences and language stacks. For instance, Haystack~\citep{haystack_get_started} offers a more production-oriented approach by treating RAG pipelines as first-class primitives. In contrast, frameworks like Semantic Kernel~\citep{semantic_kernel_agent_rag} and AutoGen~\citep{autogen_retrievechat} integrate retrieval capabilities directly into agentic workflows, with AutoGen featuring specialized agents such as RetrieveChat for complex, multi-step RAG interactions.}

\edit{For developers seeking higher-level control, DSPy~\citep{dspy_rag_tutorial} abstracts RAG pipelines into programmable modules, enabling systematic optimization of prompts and retrieval strategies. Similarly, Prompt flow~\citep{azure_promptflow_rag} facilitates the development lifecycle within cloud environments like Azure ML. The ecosystem has also expanded beyond Python, with Spring AI~\citep{spring_ai_rag} providing modular RAG and ETL support for Java-based enterprise systems, and the Vercel AI SDK~\citep{vercel_ai_sdk_rag} offering tailored solutions for TypeScript-centric, front-end applications. Collectively, these tools represent a shift from bespoke RAG implementations to standardized, reusable, and often language-agnostic development paradigms.}

\edit{\subsection{Deployment Considerations for Production RAG}
Although the RAG pattern is conceptually simple, production systems face multi-dimensional constraints, including retrieval quality, scalability, token/context budgets, latency, security/governance, and evaluation/observability~\citep{azure_rag_overview}.
In this section, we will discuss the deployment considerations for production RAG from four following aspects:}

\edit{\textbf{Scalability Challenges and Retrieval Infrastructure.}
Large-scale deployments require efficient vector indexing and ANN search.
HNSW is a widely used ANN structure with favorable scaling properties~\citep{hnsw}, and GPU-accelerated similarity search has been studied for billion-scale retrieval workloads~\citep{21tbd-faiss}.
Industry vector search systems further provide operational features (endpoints, governance, budget policies, etc.)~\citep{databricks_vector_search_docs}.}

\edit{\textbf{Retrieval Quality Optimization and System Knobs.}
Production quality improvements are typically staged and evaluation-driven.
Databricks' retrieval quality guide recommends establishing an evaluation framework first, then applying hybrid search, metadata filtering, and reranking as progressively more expensive steps~\citep{databricks_retrieval_quality_guide}.
Similarly, Azure distinguishes classic RAG and agentic retrieval for complex queries, reflecting a practical trade-off between quality and latency~\citep{azure_rag_overview, azure_agentic_retrieval}.}

\edit{\textbf{Cost-Performance Trade-Offs.}
Key cost drivers include ingestion/embedding, vector search, reranking, and LLM token usage.
Vertex AI RAG Engine explicitly itemizes billing across ingestion/parsing, embeddings, vector search (managed Spanner), and reranking, enabling cost attribution and budgeting~\citep{vertex_rag_billing}.
Token budgets further motivate careful top-$k$ selection and context compression~\citep{azure_rag_overview}.}

\edit{\textbf{Production System Requirements}
Enterprise deployments require Access Control List (ACL) aware retrieval and compliance controls~\citep{m365_copilot_retrieval_api}.
Provenance (citations) improves trust and auditability. Amazon Bedrock Knowledge Bases returns citations to specific retrieved chunks in Retrieve-and-Generate responses~\citep{aws_bedrock_kb_retrieve}.
Evaluation and monitoring are essential to prevent silent regressions; OpenAI provides an Evals API to run evaluation jobs programmatically~\citep{openai_evals_api}, and LangSmith provides observability for tracing and evaluation of LLM applications~\citep{langsmith_observability}.
Managed RAG platforms may also provide security controls such as VPC-SC and CMEK support~\citep{vertex_rag_ingestion_security}.}


%% file: 10-discussion.tex
\section{Discussion and Future Directions}
\label{sec:future}

Despite the success of the RAG for natural language processing, some challenges should be considered. 
This paper highlights these challenges to inspire future research and provides possible future research directions in RAG for NLP.

\edit{\subsection{Security and Privacy Considerations for RAG}
RAG improves factuality by grounding generation in external data sources, but it also introduces new security and privacy risks because the retrieval database (e.g., document store or vector database) becomes an additional trust boundary and attack surface. Recent studies show that manipulating or contaminating the knowledge base can directly steer downstream generation, indicating that RAG security cannot be reduced to the base LLM’s safety alone~\citep{25usenix_poisonedrag}. In practice, design choices discussed earlier, such as retrieving larger top‑$k$ sets, adopting early-fusion concatenation, or using query rewriting in query-based fusion, may amplify exposure by increasing the amount of external content injected into the model context, thereby enlarging the attack surface.}

\edit{\textbf{Privacy Risks of External Knowledge Bases.}
RAG systems typically store either raw documents, embeddings, or both. A common assumption is that embeddings are ``safe'' representations; however, research~\citep{23acl_sentenceleak} demonstrates that sentence embeddings can leak substantial information and can be inverted to recover original sentences, which challenges the privacy-by-embedding assumption. Furthermore, OWASP~\citep{owasp2025llm08} highlights that multi-tenant vector databases may suffer from cross-context leakage if access control and dataset partitioning are inadequate, allowing unauthorized retrieval across users or applications. To mitigate such risks, permission-aware retrieval (e.g., RBAC-filtered retrieval~\citep{25ns_rbac}) has been proposed to enforce least-privilege access at retrieval time, preventing sensitive passages from entering the prompt in the first place. Complementary measures can be using strong tenant isolation, auditing of retrieval logs, and designing privacy-preserving embedding defense.}

\edit{\textbf{Information Leakage Through Retrieval and Generation.}
Even when the knowledge base is not directly compromised, RAG may leak private or proprietary information through the retrieval-then-generation pipeline. 
Attackers can craft adaptive queries to induce the system to reveal sensitive sentences that originate from the knowledge base, and recent work~\citep{D25arxiv_fineprivacy} proposes black-box frameworks for fine-grained privacy extraction that explicitly identify and extract knowledge-base-derived sensitive sentences from mixed RAG outputs. 
Other studies~\citep{24arxiv_jailbreaking} further show that, under jailbreaking capabilities, attacks can be escalated toward large-scale document extraction from RAG knowledge bases, with leakage severity affected by context length and embedding configurations. 
These results imply that improving retrieval recall (e.g., larger top‑$k$) or aggressive fusion can trade off against confidentiality. 
Practical mitigations include retrieval-time authorization, sensitive-content filtering/redaction on retrieved chunks, and limiting verbatim reproduction of retrieved passages (e.g., summary-first with controlled quotation). 
Additionally, using synthetic data to replace sensitive corpora has been explored as a privacy-oriented alternative for RAG augmentation~\citep{zeng-etal-2025-mitigating}.}

\edit{\textbf{Adversarial Attacks on RAG Systems and Defenses.}
Attackers can target the knowledge base, the retriever, or the interaction between retrieval and generation via knowledge poisoning or prompt injection. 
Knowledge poisoning injects malicious texts into the knowledge base to manipulate outputs for targeted queries~\citep{25usenix_poisonedrag, chang2025one}. 
Prompt injection remains another central threat: indirect injection can be embedded in retrieved documents, blurring the line between data and instructions~\citep{owasp2025llm01}. 
Defenses can be organized into three layers: (i) ingestion-time provenance and access control, (ii) retrieval-time anomaly detection, and (iii) generation-time verification and correction. 
Representative approaches include Self-RAG~\citep{23arxiv-self-rag} that retrieves on demand with self-critique, RARR~\citep{23acl-rarr} that leverages post-hoc revision for attribution, and SelfCheckGPT~\citep{manakul-etal-2023-selfcheckgpt} that introduces sampling-based consistency checks. 
Fine-grained citations further improve auditability by grounding claims in retrieved information~\citep{xia-etal-2025-ground}.} 

\edit{\textbf{Security Evaluation for RAG.}
Given these risks, RAG evaluation should extend beyond answer accuracy and retrieval metrics to include security/privacy robustness (e.g., leakage rate, prompt-injection success rate, poisoning success rate). Benchmarks such as SafeRAG~\citep{2025acl-saferag} facilitate systematic security evaluation of RAG pipelines. For defenses, recent research explores instruction detection and removal pipelines for indirect prompt injection, but results also suggest that robustness is far from solved, motivating future work on principled trust modeling of retrieval sources, provenance-aware ingestion, and secure fusion strategies that minimize exposure while preserving utility~\citep{chen2025can}.}

\subsection{Retrieval Quality}
\edit{High-quality retrieval is fundamental to RAG systems. 
Imperfect retrieval, whether due to missing relevant passages, returning contradictory information, or introducing noise, can directly degrade generation quality and even induce hallucinations~\citep{24tacl_retriever_eval}. 
Ensuring high retrieval quality requires careful consideration of both single‑round retrieval factors (e.g., key representation, embedding models) and the key challenges in multi‑hop or iterative retrieval settings (e.g., semantic drift).}

\edit{\textbf{Improving Single-Round Retrieval Quality.} 
For each retrieval, four key factors collectively determine the quality of retrieved results. 
The first is the key choice in retriever, which should be aligned with the downstream task. 
For example, for QA tasks, adopting the question itself as the key is an intuitive yet effective choice. 
The second factor is the embedding model, which converts text into vector representations. General-purpose models like BERT~\citep{19naacl-bert} or RoBERTa~\citep{19arxiv-roberta} may suffice for broad domains, but adapting the encoder to the target domain can substantially improve semantic matching. 
Third, the similarity metric used to compare query and key embeddings plays a critical role. Beyond standard cosine or Euclidean distance~\citep{ir_similarity}, task‑specific metrics such as optimal transport distance~\citep{23nips-ra-mil} have been shown to better capture nuanced relevance in specific domain scenarios. 
Finally, the choice of ANN search algorithm (e.g., HNSW~\citep{hnsw}, IVF‑PQ~\citep{11tpami-pq}) introduces a trade‑off between search speed and accuracy; 
tuning its parameters directly affects both the precision of retrieved results and system latency.}

\edit{\textbf{Semantic Drift in Multi-hop or Iterative Retrieval.} 
Beyond single-round retrieval, many RAG systems employ multi-hop or iterative retrieval strategies to gather retrieved information incrementally. 
For example, decomposing a complex question into sub-queries, or reformulating queries based on initial retrieval results. 
While such strategies can improve coverage, they introduce the risk of \textit{semantic drift} issues~\citep{25acl-semantic-drift-1}, that is as the system retrieves multiple rounds of information, the retrieved content may gradually deviate from the original information need. 
This occurs because each retrieval step relies on the output of the previous step, and small errors or off-topic tangents can compound over iterations. 
Semantic drift might easily occur when any of the single-round retrieval factors are suboptimal. 
For example, a weakly aligned embedding model may return some irrelevant knowledge that steer subsequent queries off course. 
Consequently, mitigating semantic drift requires not only optimizing each individual retrieval step but also designing mechanisms to monitor and correct for drift, such as query validation~\citep{24acl-semantic-drift-2}, relevance feedback loops~\citep{24emnlp-rafe, 24emnlp-re-rag}, or early stopping when retrieved information becomes consistently irrelevant~\citep{23emnlp-flare, 24acl-dragin, 25arxiv-stop-rag, 24emnlp-uar}.}


\subsection{RAG Efficiency}
RAG efficiency is crucial for downstream NLP applications, which limits the volume of data that can be retrieved.
There are two simple ways to guarantee RAG efficiency without new algorithms, i.e., reducing the volume of data or adding more powerful computing and memory resources.
However, the former may impact the retrieval quality, while the latter requires more resource cost.

RAG efficiency encompasses the efficiency of the retriever and the efficiency of retrieval fusions.
Retriever efficiency refers to the time cost of retrieving relevant information, which can be divided into three parts, i.e., encoding time, ANN searching time, and data fetching time of the knowledge base. 
It is unnecessary to jointly optimize all three components as the bottleneck would vary from different database sizes.
For smaller retrieval databases, such as those with fewer than 1 million entries, the encoding phase is often the primary bottleneck, as the vector database can be all stored in the memory.
Several topics, such as model quantization~\citep{21icml-i-bert, 21acl-binarybert}, distillation~\citep{20emnlp-tinybert, 23aaai-skdbert}, or model pruning~\citep{21tacl-compress-bert}, are used to accelerate the encoding. \edit{Besides, efficient LLM serving systems (e.g., vLLM/PagedAttention) and decoding accelerators (e.g., speculative decoding) can significantly improve throughput or reduce decoding time, helping offset RAG’s longer prompts.~\citep{kwon2023efficient}}

In contrast, for larger databases, the time cost of searching in the index and fetching data from the knowledge base becomes the major bottleneck, as the searching is over a considerable amount of data, and the fetching involves I/O overheads.
In this case, efficient ANN searching algorithms~\citep{21tbd-faiss, 24arxiv-faiss, 20icml-scann} and system-level optimizations~\citep{24arxiv-ragcache, 24arxiv-piprag} are the main focus.

Retrieval fusion efficiency, which aims to enhance the inference efficiency when integrating retrievals, is worth optimizing for improving the RAG efficiency.
For example, the computational overhead of query-based fusion is often non-negligible due to the long sequence length.
Some works, such as Fid-light~\citep{23sigir-fid-light} and ReFusion~\citep{24iclr-refusion}, mainly target reducing the computations while integrating the retrieved information.

\edit{\subsection{RAG Reliability}
Reliability is another practical challenge. RAG/agentic variants may exhibit repetitive generation or ``looping'' behaviors that stall responses. Degenerative repetition has been analyzed in neural text generation and is closely related to decoding dynamics~\citep{holtzmancurious}. Production studies further report that repetition can cause severe system stalling and summarize effective countermeasures such as early-stopped beam search, decoding penalties, and training-time alignment~\citep{wang2025solving}. Decoding strategies such as contrastive search provide additional tools to reduce repetition and improve coherence~\citep{sucontrastive}. In practice, robust termination criteria (max steps/budgets, loop detectors, stop sequences), coupled with monitoring and fallback policies (e.g., reduced context, simpler answer mode), are essential for production-grade reliability.}

\edit{\subsection{RAG vs. Long-Context LLMs}
The emergence of long-context LLMs, capable of processing millions of tokens in a single forward pass, has raised a fundamental question: does RAG still matter when the model can directly inject the context of the entire knowledge base? This section examines this question from two angles. First, we compare RAG and long-context LLMs as competing paradigms, analyzing their relative strengths and weaknesses. Second, we explore how RAG can complement long-context LLMs to achieve further improvements, rather than being rendered obsolete.}

\edit{\textbf{RAG vs. Long-Context LLMs as Competing Paradigms}
Recent studies have compared the performance of RAG systems and long-context LLMs across various tasks. }
Hui et al.~\citep{24nips_ragbench} demonstrate that on free-form or knowledge-based tasks (e.g., paper-based and wiki-based QA), both paradigms perform similarly. 
However, on tasks requiring numerical reasoning (e.g., financial QA), long-context LLMs underperform compared to RAG, as the verbose content in long contexts can obscure key facts and hinder precise reasoning. 
Moreover, for smaller LLMs that struggle to process large volumes of data effectively, RAG tends to outperform long-context mechanisms. 
\edit{Beyond performance, long-context LLMs usually incur significantly higher computational overhead as input length grows, while RAG processes only selected relevant passages, offering a more economical solution.
These findings suggest that while long-context capability is valuable, it does not fully replace the need for retrieval, especially when precision, reasoning over scattered facts, resource efficiency, or costs are priorities.}

\edit{\textbf{Enhancing Long-Context LLMs with RAG.}
Rather than regarding them as mutually exclusive, RAG can be used to further improve long-context LLMs. 
Instead of feeding the entire knowledge base into the context window, a retriever selects only the most relevant information which acts as a context compressor, reducing computational cost and mitigating the ``lost in the middle'' issue~\citep{23arxiv-lost-in-middle}. 
To fully realize synergies, several improvement strategies can be adopted when applying RAG to long-context LLMs. 
For example, lightweight prompt engineering~\citep{20neurips-rag, 23acl-pgra} or calibration methods~\citep{23emnlp-flare, 23arxiv-self-rag} can offer improved results with minimal overhead; iterative RAG~\citep{23emnlp_iterag} or query refinement~\citep{24arxiv_crag} can achieve better performance and robustness.
These strategies can also be potentially combined, e.g., refined queries with iterative loops.
In essence, long-context LLMs and RAG are not competitors but complementary tools. Long context provides breadth while retrieval provides precision and verifiability.}

\subsection{RAG Training}
As introduced in Section~\ref{sec:training}, RAG training includes two branches of works, RAG with/without knowledge base update.
For RAG without a knowledge base update, the main challenge is how to jointly optimize all parameters in RAG.
This may involve new loss functions with multiple objectives, new optimizations for efficient tuning parameters in the retriever and generator, or other training strategies.

For RAG with knowledge base update, one challenge is how to align the retrieval representations with the generator's representations.
Although the time cost of the update operation in knowledge base cannot be ignored, some works~\citep{22neurips-retroprompt} reduce the update frequency by asynchronously updating, thus achieving the alignment of knowledge representation and model's representation.
Another challenge is when to retrain/fine-tune the generator in RAG when a new corpus is added.
Due to the in-context learning capability of existing LLM-based generators and high training overhead, retraining/fine-tuning the generator or directly inferring the generator becomes a challenging choice for different scenarios.
Recently, some efficient training strategies~\citep{22iclr-lora, 23nips-qlora} have been proposed to accelerate the fine-tuning process, which can be taken into consideration.

\subsection{Cross-Modality Retrieval} 
Retrieving cross-modality information in NLP tasks can greatly enhance the quality and richness of the representations, leading to improved performance. 
First, cross-modality information, such as combining text with images, videos, or audio, provides a richer context to the content~\citep{hu2023multimodal}. 
For instance, when language is ambiguous, accompanying images can clarify meanings difficult to convey through text alone. 
Second, different modalities can contribute various types of information that are not accessible from a single source. 
For example, visual data can provide spatial, color, and action cues, while textual data can offer detailed descriptions, emotions, or abstract concepts. 
Combining these can lead to a more comprehensive understanding of the data.
Moreover, Models trained on multi-modal data typically exhibit increased robustness and generalizability~\citep{multirobust}. 
These models are adept at associating information across diverse inputs, mitigating overfitting to the peculiarities of a single modality. 
This attribute is particularly valuable in real-world applications of NLP, such as in autonomous vehicles, where systems must interpret textual information from signs or dialogues and sensory data from the surrounding environment to make informed decisions.
Furthermore, multi-modal data can resolve ambiguities that cannot be resolved within a single modality. 
For example, the phrase "bank" can refer to either a financial institution or the side of a river, and visual context can help disambiguate this.
Last, human communication is inherently multi-modal, incorporating elements such as gestures, facial expressions, and tone of voice.
Systems capable of processing multiple modes of communication can interact with humans in a manner that is both more natural and intuitive.
In conclusion, integrating cross-modality information in RAG for NLP tasks not only enhances the richness and quality of data representations but also significantly improves the systems' comprehension, interaction capabilities, and adaptability to diverse applications.

\edit{\subsection{GraphRAG and Graph-based Retrieval}
Recent open-source and research advances have pushed RAG beyond flat passage retrieval toward graph-based retrieval, where documents are transformed into structured entity–relation representations and retrieval operates over graph neighborhoods or communities. A prominent example is GraphRAG, which uses an LLM to build a graph index in two stages: (i) constructing an entity knowledge graph from source documents, and (ii) pre-generating community summaries for groups of closely related entities. At query time, community summaries are used to produce partial answers that are then aggregated into a final response~\citep{edge2024local}.
Microsoft has also released an open-source GraphRAG implementation, enabling practitioners to reproduce and extend graph-based pipelines, while noting that graph indexing can be operationally expensive~\footnote{https://github.com/microsoft/graphrag}.
At the same time, recent analyses emphasize that graph-based RAG does not uniformly dominate “vanilla” RAG. The benefits depend on the task structure (e.g., hierarchical or multi-hop reasoning) and may entail higher latency and graph construction overhead, motivating careful scenario-based evaluation and benchmarking~\citep{xiang2025use}.}

%% file: 11-conclusion.tex
den\section{Conclusion}
\label{sec:con}
\edit{In this survey, we presented a systematic review of RAG for NLP. 
We introduced its core components, including retriever, generator, and retrieval fusions, and provided a novel taxonomy of fusion methods with structured comparisons across accessibility, efficiency, and use cases. 
We further examined RAG applications across diverse NLP tasks and discussed evaluation methodologies and benchmark limitations. 
Training paradigms with and without knowledge base updates were also analyzed. 
Finally, we explored industrial deployment considerations and identified emerging challenges and future directions, such as security or graph-based retrieval. 
We hope this survey serves as a practical guide for researchers and practitioners advancing RAG systems.}

%% file: 12-declare.tex
\section{Statements and Declarations}

\subsection{Authors' Contribution}
S.W. and Y.X., wrote and significantly revised the main manuscript. 
Y.C., H.W., C.C., and Y.Y. drafted several sections. 
S.W., Y.X., Y.C., H.W., C.C., Y.Y, L.H, X.L, T.K, N.G, and C.X participated in the main investigation and the design of the paper’s structure.
All authors participated in designing and refining the methodology.
All authors reviewed the manuscript.

\subsection{Conflict of Interest}

 The authors declare no Conflict of interest.